\documentclass[10pt,journal,compsoc]{IEEEtran}
\usepackage{amsmath,amsfonts}
\usepackage{algorithm}
\usepackage{algpseudocode}
\usepackage{array}
\usepackage[caption=false,font=footnotesize,labelfont=sf,textfont=sf]{subfig}
\usepackage{textcomp}
\usepackage{stfloats}
\usepackage{url}
\usepackage{verbatim}
\usepackage{graphicx}
\usepackage{hyperref}
\usepackage[figure]{hypcap}
\usepackage{booktabs}
\usepackage{bm}
\usepackage{pifont}
\usepackage{xcolor}
\usepackage{colortbl}
\usepackage{multirow}
\usepackage{amssymb}
\usepackage{array}
\usepackage{colortbl}
\usepackage{hhline}
\newcolumntype{P}[1]{>{\centering\arraybackslash}p{#1}}

\usepackage{pgfplots}
\usepgfplotslibrary{groupplots}
\pgfplotsset{compat=1.16}

\def\BibTeX{{\rm B\kern-.05em{\sc i\kern-.025em b}\kern-.08em
    T\kern-.1667em\lower.7ex\hbox{E}\kern-.125emX}}
\usepackage{balance}
\begin{document}
\title{TINA+: Probing Residual Visual Knowledge in Unlearned Diffusion Models via Diffusion-Consistent Text-Free Inversion}
\author{Qianlong Xiang, Miao Zhang$^{\dagger}$,~\IEEEmembership{Member, IEEE}, Kun Wang, Haoyu Zhang, Junhui Hou$^{\dagger}$,~\IEEEmembership{Senior Member, IEEE}, Liqiang Nie,~\IEEEmembership{Senior Member, IEEE}
\IEEEcompsocitemizethanks{
    \IEEEcompsocthanksitem ${\dagger}$ Corresponding author.
    \IEEEcompsocthanksitem Qianlong Xiang is with the School of Computer Science and Technology, Harbin Institute of Technology (Shenzhen), Shenzhen 518055, China, the City University of Hong Kong, Hong Kong SAR, China, and the Shenzhen Loop Area Institute, Shenzhen 518045, China (e-mail: xiangqianlongcs@gmail.com).
    \IEEEcompsocthanksitem Miao Zhang and Liqiang Nie are with the Harbin Institute of Technology (Shenzhen), Shenzhen 518055, China (e-mails: zhangmiao@hit.edu.cn and nieliqiang@gmail.com).
    \IEEEcompsocthanksitem Kun Wang is with the National University of Singapore, Singapore (e-mail: khylon.kun.wang@gmail.com).
    \IEEEcompsocthanksitem Haoyu Zhang is with the Harbin Institute of Technology (Shenzhen), Shenzhen 518055, China  and Pengcheng Laboratory, Shenzhen 518000, China (e-mail: zhang.hy.2019@gmail.com).
    \IEEEcompsocthanksitem Junhui Hou is with the Department of Computer Science, City University of Hong Kong, Hong Kong, China (e-mail: jh.hou@cityu.edu.hk).
}
}

\definecolor{qianlong}{RGB}{0,0,0}
\newcommand{\xql}[1]{{#1}}

\IEEEtitleabstractindextext{
\begin{abstract}
Although text-to-image diffusion models exhibit remarkable generative power, concept erasure techniques are essential for their safe deployment to prevent the creation of harmful content.
\xql{
To evaluate these erasure methods, a series of adversarial probes have been designed to test whether erased concepts can still be recovered, in turn driving the development of stronger erasure defenses.
However, existing erasure and probe methods remain largely confined to a text-centric paradigm, focusing on whether the text-to-image mapping is severed while overlooking whether the corresponding visual knowledge still remains.
To investigate this overlooked question, we adopt a visual perspective and leverage diffusion inversion to probe whether a generative trajectory can still be found to reconstruct visual instances of the erased concept.
A natural starting point is standard diffusion inversion, where the text prompt facilitates faithful reconstruction.
However, this textual condition is precisely what text-centric defenses suppress, and its involvement would also prevent a purely visual assessment of residual knowledge.
Operating instead under a null-text condition removes this dependence and enables a purely visual probe.
This benefit comes at a cost: the absence of textual conditioning amplifies the minor approximation errors inherent in standard diffusion inversion, hindering faithful trajectory recovery.
To address these challenges, we introduce TINA+, a diffusion-consistent Text-free INversion Attack equipped with an optimization-based inversion procedure to improve null-text inversion accuracy.
Beyond inversion accuracy, we find that unconstrained diffusion inversion may discover spurious trajectories, even allowing a randomly initialized diffusion model to reconstruct the target concept.
Such trajectories are inconsistent with the diffusion process and may falsely indicate the presence of residual visual knowledge.
TINA+ therefore introduces Diffusion-Consistent Trajectory Regularization to suppress this failure mode.
By penalizing trajectories that fall far below the expected marginal energy evolution of diffusion, TINA+ avoids spurious inversion paths while preserving its ability to recover erased concepts through diffusion-consistent trajectories.
Extensive experiments across twelve erasure methods, four concept-erasure tasks, and different model architectures demonstrate that TINA+ reliably probes residual visual knowledge through diffusion-consistent visual trajectories.
These results provide stronger evidence that current methods often obscure concepts by severing text-image links, rather than eliminating the underlying visual knowledge.
The project page is \url{https://qianlong0502.github.io/TINA-Plus-Homepage/}.
}
\end{abstract}

}

\maketitle

\newcolumntype{P}[1]{>{\centering\arraybackslash}p{#1}}

\section{Introduction}
\label{sec:intro}
With the rapid development of deep learning \cite{vaswani2017attention,zhang2023attribute,li2026optimus3,wang2025unifying,wang2025decad,liu2025understanding,wang2026cross,wang2024explicit,wang2025redundancy,ntire26visage,guan2026spaceerapp}, generative artificial intelligence, particularly Text-to-Image (T2I) diffusion models such as Stable Diffusion is fundamentally reshaping the landscape of digital content creation~\cite{rombach2022high,nichol2021glide,ramesh2021zero,saharia2022photorealistic,xiang2025dkdm,chen2023pixart,peebles2023scalable,ju2023pnp,miyake2025negative,mokady2023null,zhang2025easyinv}.
Their remarkable capacity for creative synthesis has catalyzed countless applications, from art and design to entertainment and media.
However, since these models are typically trained on massive, unfiltered datasets scraped from the Internet~\cite{schuhmann2022laion}, their proliferation brings a turbulent undercurrent of ethical and safety challenges.
The same models that generate breathtaking art can be misused to create deepfakes that violate personal privacy~\cite{carlini2023extracting}, imitate artistic styles that infringe on copyright~\cite{jiang2023ai,roose2022ai,somepalli2023diffusion}, and produce harmful Not-Safe-For-Work (NSFW) imagery~\cite{zhang2024generate,hunter2023ai,schramowski2023safe}, posing risks to intellectual property and societal well-being.\begin{figure}
  \centering
  \includegraphics[width=\linewidth]{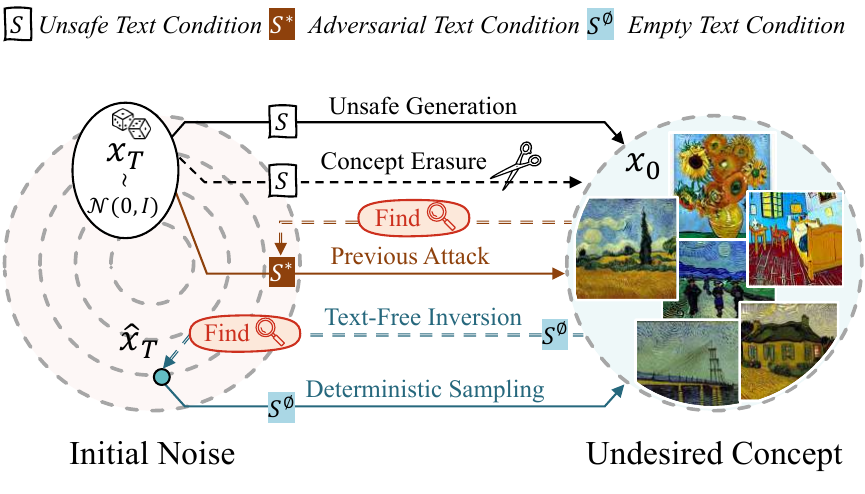}
  \caption{
  \xql{Conceptual overview of text-centric erasure vulnerabilities and our TINA+ attack.}
  \textbf{Concept Erasure} usually severs the link between a specific text condition and the undesired concept.
  \textcolor[RGB]{142, 65, 12}{\textbf{Previous Attacks}} remain text-centric, finding adversarial text condition to reactivate the concept.
  \xql{\textcolor[RGB]{38, 104, 125}{\textbf{Our TINA+}} bypasses the text pathway entirely. Using an empty text condition, it finds a noise that regenerates the concept through a diffusion-consistent trajectory, providing stronger evidence that residual visual knowledge persists in existing erased models.}
  }
  \label{fig:intro}
\end{figure}

To address these challenges, the field has converged on the paradigm of \textit{Concept Erasure}, a specialized form of \textit{Machine Unlearning}~\cite{xu2023machine}.
This paradigm aims to mitigate risks by directly modifying model parameters.
For instance, approaches such as Erasing Stable Diffusion (ESD) and Ablating Concepts (AC) retrain specific model layers to remap a forbidden concept representation to that of a neutral or null concept, thereby severing the explicit link between a text prompt and its corresponding visual output~\cite{gandikota2023erasing,kumari2023ablating}.
Other methods like Forget-Me-Not (FMN) minimize the attention maps corresponding to the target concept~\cite{zhang2024forget}, while Unified Concept Editing (UCE) applies efficient, closed-form edits to the cross-attention weights~\cite{gandikota2024unified}.
Despite their distinct mechanisms, these approaches share a common, text-centric operational principle, equating erasure with merely severing its \textit{text-image mapping}.

The text-centric paradigm, as conceptualized in Figure~\ref{fig:intro}, has consequently guided adversarial attacks almost exclusively to the textual domain.
Initial red-teaming efforts involve prompt-level manipulations, attempting to find alternative phrasings or complex prompts that reactivate the target concept~\cite{chin2024prompting4debugging,zhang2024generate,tsai2024ring}.
Moreover, the more sophisticated embedding-space attacks, which move beyond discrete tokens, remain fundamentally dependent on the textual conditioning pathway~\cite{pham2024circumventing}.
Generally, these red-teaming methods aim to discover new token embeddings~\cite{pham2024circumventing} or optimize an adversarial text prompt~\cite{zhang2024generate} that serves as a proxy for the erased concept.
In response to this threat, an interplay between text-based attacks and defenses has driven the subsequent development of more resilient safeguards, such as AdvUnlearn~\cite{zhang2024defensive} and STEREO~\cite{srivatsan2025stereo}.
This adversarial co-evolution has produced defenses that demonstrate increasing efficacy, thereby fostering an appearance of robust model security.

However, in this work, we contend that this competitive co-evolution of text-centric methods rests on a fundamental and fatal assumption.
Specifically, existing paradigm mistakenly equates the erasure of a \textit{text-to-image link} with the far more complex task of eliminating the underlying \textit{visual knowledge} from the parameter space.
Consequently, existing methods critically ignore the fact that even if textual inputs can no longer induce the target concept, the corresponding visual knowledge may still persist, untapped, within the model.
To challenge this dominant paradigm and substantiate our claim, we introduce a core hypothesis:

\xql{\textit{If concept erasure primarily severs text-image mappings, the underlying visual knowledge may still persist within the erased model, enabling the erased concept to be regenerated through corresponding text-free generative trajectories.}}

\xql{To validate this hypothesis from a purely visual perspective, thereby challenging the efficacy of these text-centric defenses, we introduce TINA+, a diffusion-consistent Text-free INversion Attack.}
\xql{Unlike prior embedding-space attacks that seek a new textual proxy for the erased concept~\cite{pham2024circumventing}, TINA+ is designed to completely bypass the textual conditioning pathway.}
\xql{To achieve this, TINA+ leverages Denoising Diffusion Implicit Models (DDIM)~\cite{song2020denoising} inversion as a visual probe (illustrated in Figure~\ref{fig:intro}), allowing us to investigate the visual knowledge embedded within the model directly.}
\xql{However, identifying such a generative trajectory is challenging. Standard diffusion inversion typically relies on textual guidance to facilitate faithful reconstruction, but this guidance is precisely what text-centric defenses suppress and would also prevent a purely visual assessment of residual knowledge.}
\xql{Operating under the null-text condition necessary for a purely visual probe removes this dependence, but amplifies the minor approximation errors inherent in standard inversion.}
\xql{To address this imprecision, TINA+ incorporates an optimization procedure that corrects for these errors, allowing it to more accurately map a target image back to a latent noise vector.}
\xql{When used to initialize the subsequent generation process, the resulting latent vector can lead an erased model to regenerate the forbidden content through a text-free generative trajectory, entirely bypassing the text-conditioning mechanism.}

\xql{
However, accurate text-free inversion alone is not sufficient to provide a reliable probe of residual visual knowledge. An optimization-based inversion procedure may overfit the target image without necessarily reflecting a plausible diffusion generation process. In extreme cases, even a randomly initialized diffusion model, which contains no meaningful learned visual knowledge of the target concept, can be inverted to reconstruct the target concept through a spurious, diffusion-inconsistent trajectory. This observation reveals an important limitation: visual resemblance alone does not necessarily certify that the erased model truly preserves the corresponding concept knowledge. A reliable text-free probe should therefore not only regenerate the target concept, but also ensure that the discovered trajectory remains compatible with the expected marginal energy behavior of diffusion. To address this issue, TINA+ introduces Diffusion-Consistent Trajectory Regularization. Specifically, TINA+ penalizes trajectories whose latent energy falls far below the expected marginal energy evolution of diffusion, thereby suppressing severe energy collapse along spurious inversion paths. We further introduce a forward marginal initialization strategy to bypass the fragile low-noise region and stabilize trajectory discovery. With these designs, TINA+ serves not merely as a reconstruction-oriented attack, but as a diffusion-consistent visual probe that better distinguishes genuine residual visual knowledge from artifacts caused by unconstrained inversion.
}

\xql{
Extensive experiments across four concept-erasure tasks evaluate TINA+ against twelve state-of-the-art erasure methods and six representative attacks. The results show that TINA+ remains effective even when existing text-centric attacks are substantially suppressed, supporting the persistence of text-independent visual trajectories in erased models. Negative-control experiments and trajectory diagnostics further demonstrate that the proposed diffusion-consistency constraints suppress spurious reconstructions, making TINA+ a more reliable probe of residual visual knowledge. Additional latent embedding and architecture-generalization analyses show that the discovered trajectories activate concept-specific internal representations and that the exposed vulnerability is not limited to a particular diffusion architecture. Together, these findings provide compelling evidence that current erasure methods often obscure concepts by severing text-image links rather than fully removing the underlying visual knowledge, highlighting the need for more robust unlearning paradigms that operate directly on internal visual representations within T2I models.
}

To sum up, our contributions are as follows.

\begin{itemize}
  \item To our knowledge, we are the first to identify a fundamental limitation of the text-centric concept erasure paradigm: severing text-image mappings does not necessarily eliminate the underlying visual knowledge from diffusion models.
  \item \xql{We propose TINA+, a text-free attack that performs optimization-based null-text inversion under diffusion-consistent trajectory regularization to reliably probe residual visual knowledge in erased models and assess the effectiveness of concept erasure.}
  \item \xql{Extensive experiments across twelve erasure methods, four concept-erasure tasks, and different model architectures, together with random-model controls and latent embedding analysis, demonstrate the effectiveness, reliability, and generalizability of TINA+.}
\end{itemize}

\xql{
The preliminary conference version of this work introduced TINA, an optimization-based text-free inversion attack, at CVPR 2026~\cite{xiang2026tina}. This journal version develops it into TINA+ and offers three major improvements:
\begin{itemize}
  \item We provide a deeper analysis of text-free inversion and identify a previously overlooked failure mode: optimizing inversion accuracy alone can lead to spurious, diffusion-inconsistent trajectories, which may even produce visually recognizable target-like images on a randomly initialized diffusion model. This motivates a more reliable notion of trajectory validity beyond visual reconstruction.
  \item We extend the original TINA framework to TINA+ by introducing diffusion-consistent trajectory regularization. Specifically, we derive a marginal energy constraint from the forward diffusion marginal to penalize severe latent energy collapse, and further introduce forward marginal initialization to stabilize trajectory discovery in the fragile low-noise region.
  \item We substantially expand the experimental evaluation to validate the proposed diffusion-consistent probing framework. In addition to the original attack success evaluation, we include trajectory-level diagnostics, validity-aware analysis, ablation studies of the new components, and additional robustness studies to demonstrate that TINA+ can better distinguish genuine residual visual knowledge from artifacts caused by unconstrained inversion.
\end{itemize}
}

\section{Related Work}
\label{sec:related}

\subsection{Text-to-Image Diffusion Models}
The advent of Text-to-Image (T2I) diffusion models, such as Stable Diffusion~\cite{rombach2022high}, DALL-E 2~\cite{ramesh2022hierarchical}, and Imagen~\cite{saharia2022photorealistic}, has marked a paradigm shift in digital content generation. These models operate through a two-stage process: a forward diffusion process that incrementally adds noise to an image, and a learned reverse process that denoises a random vector back into a coherent image~\cite{ho2020denoising}. To achieve computational efficiency, models like Stable Diffusion perform this process in a lower-dimensional latent space~\cite{rombach2022high}. Critically, their generative process is guided by textual prompts, typically encoded and injected into the denoising network via a cross-attention mechanism~\cite{vaswani2017attention}. This tight coupling between text and image synthesis is the source of their remarkable controllability, but also the root of significant safety and ethical concerns, as the models can be prompted to produce NSFW imagery~\cite{zhang2024generate,schramowski2023safe}, imitate copyrighted styles~\cite{jiang2023ai,somepalli2023diffusion}, or violate personal privacy~\cite{carlini2023extracting}.

\subsection{Concept Erasure in Diffusion Models}
To address the misuse of T2I models, concept erasure has emerged as a critical field.
While methods like SLD~\cite{schramowski2023safe} focus on inference-time interventions, the dominant paradigm involves fine-tuning-based methods that modify model parameters.
This paradigm originated with foundational works like ESD~\cite{gandikota2023erasing} and AC~\cite{kumari2023ablating}, which retrained models to break associations with specific text prompts.
Subsequently, this was followed by more efficient and targeted techniques, such as FMN~\cite{zhang2024forget}, UCE~\cite{gandikota2024unified}, MACE~\cite{lu2024mace}, and the adapter-based SPM~\cite{lyu2024one}, often focusing on cross-attention layers.
\xql{Concurrently, general unlearning frameworks like SalUn~\cite{fan2024salun}, Scissorhands (SH)~\cite{wu2024scissorhands}, and EraseDiff~\cite{wu2025erasing} were proposed for the concept erasure task.}
As text-based attacks began to bypass these initial erasure methods, advanced adversarially robust methods were developed, including RECE~\cite{gong2024reliable}, AdvUnlearn~\cite{zhang2024defensive}, and STEREO~\cite{srivatsan2025stereo}.
These methods usually integrate attack considerations into their design to defend against adversarial prompts and embeddings.
Ultimately, this co-evolution of text-based attacks and defenses has solidified a dominant, text-centric paradigm: most of them operate by identifying or severing the \textit{text-to-image mapping}, rather than addressing the underlying visual knowledge.

\subsection{Adversarial Attacks on Concept Erasure}
The prevailing text-centric assumption of erasure methods has consequently shaped the landscape of adversarial attacks, which primarily seek to circumvent these textual defenses.
These attacks can be classified by their operational domain.
\xql{A broad class of attacks operates through textual prompts by searching for or optimizing discrete token sequences that trigger the erased concept, including PEZ~\cite{wen2023hard}, P4D~\cite{chin2024prompting4debugging}, RAB~\cite{tsai2024ring}, MMA~\cite{yang2024mma}, and UDA~\cite{zhang2024generate}.}
A more sophisticated class of attacks operates in the continuous embedding space.
These methods aim to find a novel token embedding that acts as a proxy for the forbidden concept, effectively creating a new activation pathway to the visual knowledge.
For instance, CCE~\cite{pham2024circumventing} leverages textual inversion to discover surrogate embeddings for erased concepts.

Critically, despite their varying levels of sophistication, from simple phrasings to optimized vectors, these existing attacks share a fundamental limitation: most of them presuppose that the generative capabilities of the model must be accessed via its text-conditioning pathway.
They focus on finding a new textual or embedding-space vector to bypass modified controls.
\xql{In contrast, TINA+ targets a fundamentally different attack surface by probing whether visual knowledge persists independently of textual control.
It bypasses the text-conditioning mechanism and discovers diffusion-consistent, text-free generative trajectories, enabling a reliable assessment of whether the erased model still retains the corresponding visual knowledge.}

\section{Preliminaries}
\label{sec:preliminaries}

\subsection{Latent Diffusion Models.}
Latent Diffusion Models (LDMs)~\cite{rombach2022high} generate images by reversing a $T$-step diffusion process in a compressed latent space. An image $x$ is first encoded into its latent $z_0 = \mathcal{E}(x)$ by a VAE encoder. The forward process creates a sequence of noised latents $\{z_1, \dots, z_T\}$.
A denoising U-Net $\epsilon_{\theta}$ learns to reverse this process. At each timestep $t$, it predicts the noise $\epsilon$ in $z_t$, guided by a textual condition $c$. This condition is produced by a text encoder $\mathcal{T}_{\text{text}}$ from a prompt $y$ and integrated via cross-attention. The training objective minimizes the noise prediction error:
\begin{equation}
\mathcal{L} = \mathbb{E}_{z_0, y, \epsilon, t} \left[ \left\| \epsilon - \epsilon_{\theta}(z_t, t, c) \right\|_2^2 \right].
\end{equation}
This objective trains $\epsilon_{\theta}$ to denoise $z_t$ back towards $z_0$ in a manner faithful to the condition $c$. The final image is retrieved using the VAE decoder $\mathcal{D}$.
Following the DDPM \cite{ho2020denoising} convention, we denote the per-step noise variance by $\beta_t$, the corresponding signal coefficient by $\alpha_t=1-\beta_t$, and the cumulative signal coefficient by
\begin{equation}
\bar{\alpha}_t=\prod_{s=1}^{t}\alpha_s .
\end{equation}
Throughout this paper, we use $\bar{\alpha}_t$ for the cumulative coefficient that appears in the forward diffusion marginal.
\subsection{DDIM Sampling.}
Denoising Diffusion Implicit Models (DDIM)~\cite{song2020denoising} enable deterministic generative sampling from LDMs by using a non-Markovian reverse process with zero sampling variance. Consequently, for a fixed model $\theta$ and condition $c$, any initial noise $z_T \sim \mathcal{N}(0, \mathbf{I})$ is deterministically mapped to a generated latent $z_0$.

This deterministic path is defined by an iterative update rule. Given the latent $z_t$, the model first predicts the clean latent $\hat{z}_0(z_t)$:
\begin{equation}
\label{eq:ddim_pred_z0}
\hat{z}_0(z_t) = \frac{z_t - \sqrt{1-\bar{\alpha}_t}\epsilon_\theta(z_t, t, c)}{\sqrt{\bar{\alpha}_t}}.
\end{equation}
The preceding latent $z_{t-1}$ is then computed as:
\begin{equation}
\label{eq:ddim_sample}
z_{t-1} = \sqrt{\bar{\alpha}_{t-1}} \hat{z}_0(z_t) + \sqrt{1-\bar{\alpha}_{t-1}} \cdot \epsilon_\theta(z_t, t, c),
\end{equation}
where $\{\bar{\alpha}_t\}_{t=0}^T$ are cumulative signal coefficients determined by the fixed noise schedule, with $\bar{\alpha}_0=1$.
\subsection{DDIM Inversion.}
Building upon the deterministic nature of DDIM sampling, DDIM inversion provides a mechanism to reverse the generative process. Its objective is to map a given image, represented by its initial latent $z_0$, back to a seed latent $z_T$ from which the image can be reproduced through sampling. This capability is crucial for applications like real image editing and analysis.

Ideally, the inversion would be a perfect algebraic reversal of the sampling step in Eq.~(\ref{eq:ddim_sample}), which can be expressed as:
\begin{equation}
\label{eq:ddim_inversion_ideal}
z_t = C_1(t) z_{t-1} + C_2(t) \cdot \epsilon_\theta(z_t, t, c),
\end{equation}
where 
\begin{gather}
C_1(t) = \frac{\sqrt{\bar{\alpha}_t}}{\sqrt{\bar{\alpha}_{t-1}}}, \label{eq:c1_coeff} \\
C_2(t) = \sqrt{1-\bar{\alpha}_t} - \sqrt{\frac{\bar{\alpha}_t(1-\bar{\alpha}_{t-1})}{\bar{\alpha}_{t-1}}}. \label{eq:c2_coeff}
\end{gather}
However, a practical challenge emerges: computing the latent $z_t$ requires a noise prediction $\epsilon_\theta(z_t, t, c)$, which itself depends on $z_t$, as shown in Eq.~(\ref{eq:ddim_inversion_ideal}). To break this circular dependency, standard DDIM inversion employs a key approximation: it estimates the required noise using the latent from the previous step, $z_{t-1}$, at timestep $t-1$. This results in the following iterative formula to approximate $z_t$ from $z_{t-1}$:
\begin{equation}
\label{eq:ddim_inversion_approx}
z_t \approx f_{\theta}(z_{t-1}, t, c),
\end{equation}
where
\begin{equation}
\label{eq:ddim_inversion_approx_f}
f_{\theta}(z_{t-1}, t, c)=
C_1(t) z_{t-1} + C_2(t) \cdot \epsilon_\theta(z_{t-1}, t-1, c).
\end{equation}
Starting from the clean image latent $z_0$, this equation is applied sequentially for $t=1, \dots, T$ to trace the trajectory back to an approximate seed latent $\hat{z}_T$. This vector serves as an estimate of the desired seed in the high-noise latent space required for reconstructing the original image.

\begin{figure*}[t]
    \centering
    \includegraphics[width=\linewidth]{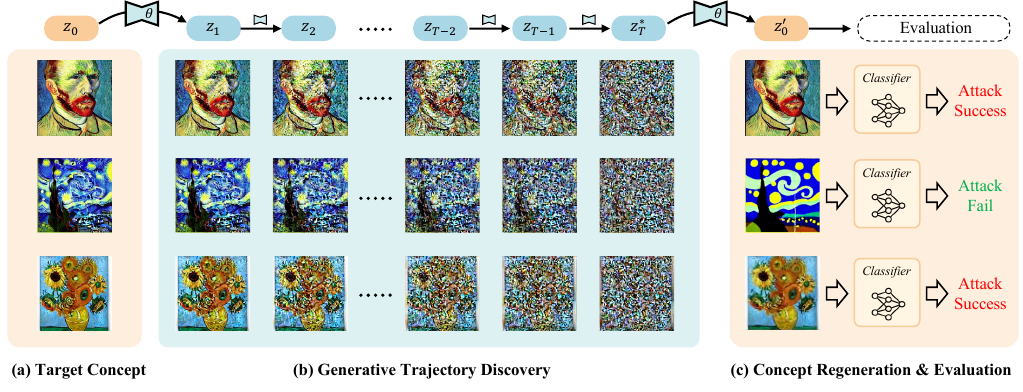}
    \caption{\xql{Framework overview of our proposed attack. Given target images containing the erased concept, we identify generative trajectories and obtain the corresponding seed latents. These seed latents are then used to generate new images with the same concept-erased model. If the generated images still contain the concept that should have been erased, the model is considered to retain the corresponding visual knowledge.}}
    \label{fig:framework_overview}
\end{figure*}

\begin{figure*}[t]
  \centering
  \includegraphics[width=\linewidth]{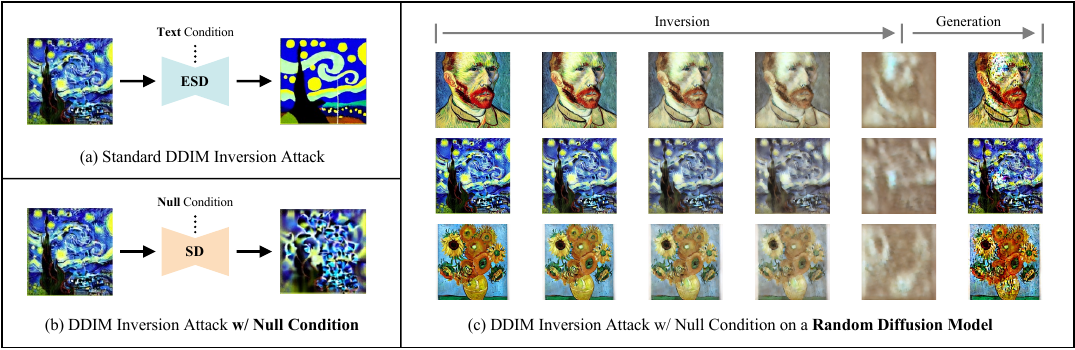}
  \caption{\xql{Standard DDIM inversion is insufficient for reliable visual trajectory discovery. (a) Text-conditioned inversion is resisted by the concept-erased model, since the corresponding text-to-image mapping has been suppressed. (b) Null-text inversion avoids the textual defense but suffers from accumulated inversion errors, failing to faithfully recover the target concept even on the unerased model. (c) Null-text inversion can even produce visually recognizable target-like images on a randomly initialized diffusion model through an unnatural, diffusion-inconsistent trajectory.}}
  \label{fig:ddim_inv_fail}
\end{figure*}

\section{\xql{Framework Overview}}
\label{sec:framework_overview}

\xql{
Figure~\ref{fig:framework_overview} provides a high-level overview of our attack framework.
Given a concept-erased diffusion model $\epsilon_{\theta}$ and a target image $x$ containing the erased concept, our goal is not to search for an adversarial textual prompt, but to examine the model from the visual latent space and determine whether it still admits a generative pathway for this concept.
We first encode the target image into the latent space as $z_0=\mathcal{E}(x)$.
The attack then identifies an inverted seed latent, denoted as $z_T^*$, by tracing the target latent toward the high-noise end of the diffusion process.
This visual trajectory discovery stage provides a visual-space probe for examining whether the erased model still retains generative knowledge of the target concept.
The seed latent $z_T^*$ serves as the starting point for sampling: when the erased model starts from this seed, it is expected to traverse a corresponding reverse trajectory and synthesize content associated with the target concept.
}

\xql{
Once the seed latent $z_T^*$ is obtained, we feed it back into the same concept-erased model $\epsilon_{\theta}$ and perform sampling to generate a new latent $z_0'$, which is decoded into an image $x'=\mathcal{D}(z_0')$.
This regeneration stage closes the loop of our visual probe: it tests whether the seed latent discovered from the target concept can drive the erased model to synthesize the corresponding concept again.
Finally, we evaluate the generated image using concept classifiers to determine whether the concept that should have been erased is regenerated.
If the generated image is recognized as containing the erased concept, the attack is considered successful. Otherwise, no residual knowledge is detected under this visual-space probe.
}

\xql{
Generally, given a set of images representing a target concept, our framework discovers visual generative trajectories associated with these images and then tests whether the discovered seed latents can drive the erased model to generate the supposedly erased concept.
However, successful regeneration alone is insufficient evidence of residual visual knowledge, because it may arise from a spurious trajectory that is inconsistent with the diffusion generation process.
The following sections detail how such visual trajectories are identified, optimized, and regularized to reveal whether a concept-erased model still preserves a generative pathway to the erased concept.
}

\section{Diffusion Inversion for Visual Trajectory Discovery}
\label{sec:diffusion_inversion_discovery}

\xql{
The framework above reduces our attack to a visual trajectory discovery problem: given a target latent $z_0$ associated with the erased concept, we aim to find a seed latent $z_T^*$ that can drive the erased model to synthesize the target concept through sampling.
Diffusion inversion provides a natural way to instantiate this trajectory discovery.
In particular, DDIM inversion traces a given image latent back toward the high-noise seed latent space by reversing the sampling process, making it a representative tool for discovering the generative trajectory associated with a target image.
If such an inversion succeeds on a concept-erased model, the resulting seed latent can be used to test whether the model still retains a visual pathway to the erased concept.
}

However, \xql{directly applying standard DDIM inversion to concept-erased models} is fundamentally flawed and exposes several important limitations, as illustrated in Figure~\ref{fig:ddim_inv_fail}.
First, if the inversion is attempted using the text condition $c$ of the target concept (e.g., ``Van Gogh style''), the \xql{concept-erased model} $\epsilon_{\theta}$ actively counteracts this prompt, leading to a failure in reconstructing the erased concept.
This confirms that text-centric defenses are indeed working against text-guided processes.
\xql{
In other words, text-guided inversion still relies on the very text-to-image mapping that concept erasure aims to suppress, and therefore cannot serve as a reliable tool for probing text-independent visual knowledge.
As shown in Figure~\ref{fig:ddim_inv_fail}(a), such a text-conditioned inversion process is still trapped in the same text-centric interaction between erasure and attack, rather than providing a purely visual exploration of the erased model.
}

Second, we could attempt the inversion under a null-text condition ($c = c_{\mathrm{null}}$) to bypass this textual defense.
This approach also fails, but in a more subtle way.
As shown in Figure~\ref{fig:ddim_inv_fail}(b), the resulting image is visually closer to the original than the text-conditioned attempt, yet it still fails to faithfully regenerate the specific target concept.
This partial failure stems from the standard inversion formula (Eq.~(\ref{eq:ddim_inversion_approx})), which relies on a critical approximation: it uses the noise prediction from the previous step, $\epsilon_{\theta}(z_{t-1}, t-1, c)$, to estimate the next latent $z_t$.
This approximation is heavily dependent on a meaningful text condition $c$ to guide the noise prediction.
In the absence of such guidance ($c=c_{\mathrm{null}}$), the predictions become unconstrained, and the small error introduced at each step rapidly compounds across the entire trajectory.
Consequently, the final computed seed latent $\hat{z}_T$ drifts sufficiently from the desired seed latent $z_T^*$ to lose the high-fidelity details of the target concept, rendering it incapable of accurately generating the target concept again.

\begin{figure*}[t]
  \centering
  \includegraphics[width=\linewidth]{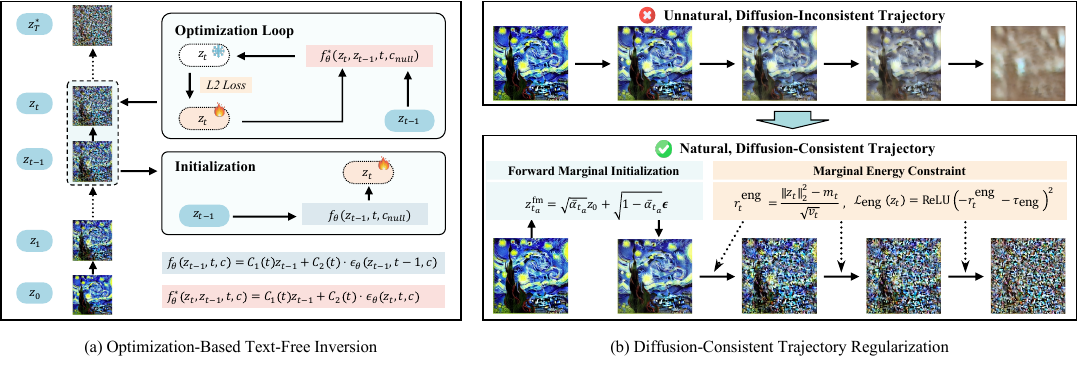}
  \caption{\xql{Overview of TINA+. Given a target image, we encode it into a latent $z_0$ and perform text-free inversion under the null-text condition. TINA+ first refines each intermediate latent through fixed-point consistency to correct accumulated inversion errors. Furthermore, TINA+ applies diffusion-consistent trajectory regularization, including forward marginal initialization and marginal energy loss, to avoid spurious diffusion-inconsistent trajectories. The resulting seed latent $z_T^*$ is then used for generation with the same concept-erased model.}}
  \label{fig:method}
\end{figure*}

\xql{
More importantly, we further observe a more severe limitation when applying null-text DDIM inversion to a completely randomly initialized diffusion model.
As shown in Figure~\ref{fig:ddim_inv_fail}(c), even though the random model contains no meaningful generative knowledge of the target concept, standard inversion can still trace the target image toward a seed latent and then produce an image that preserves certain coarse visual structures of the input.
However, the intermediate trajectory is highly unnatural from the perspective of diffusion dynamics: it is diffusion-inconsistent and deviates substantially from the typical progression of a diffusion generation process.
This observation suggests that visual resemblance produced by unconstrained inversion does not necessarily indicate that the model truly retains the corresponding concept knowledge.
In other words, if even a random diffusion model can yield an apparently concept-related trajectory, then standard DDIM inversion alone is insufficient for evaluating whether a concept-erased model actually preserves the erased visual knowledge.
}

\xql{
These observations show that diffusion inversion is a natural starting point for visual trajectory discovery, but standard inversion is insufficient as a reliable trajectory-discovery method for attacking concept-erased models.
The key challenge is therefore not merely to invert a target image, but to identify a seed latent whose generation trajectory remains faithful to the target concept under the erased model while avoiding spurious, diffusion-inconsistent trajectories that can arise from unconstrained inversion.
Motivated by this challenge, we introduce TINA+, an optimization-based text-free inversion method that refines the intermediate latents along the trajectory to correct the accumulated approximation errors of standard inversion.
Furthermore, TINA+ regularizes the discovered trajectory with diffusion-consistency constraints, so that successful regeneration is supported by a trajectory that is more consistent with the diffusion generation process rather than by a spurious inversion path.
}

\section{Diffusion-Consistent Text-Free Inversion}
\label{sec:method}

\xql{
The previous section shows that standard DDIM inversion is insufficient for reliable visual trajectory discovery.
In particular, text-conditioned inversion is resisted by the erased text-to-image mapping, while null-text inversion avoids the textual pathway but introduces two remaining challenges: inaccurate inversion caused by accumulated approximation errors, and unnatural inversion caused by spurious trajectories.
As illustrated in Figure~\ref{fig:method}, TINA+ addresses these challenges through optimization-based text-free inversion and diffusion-consistent trajectory regularization.
The former enforces fixed-point consistency to correct inversion errors, while the latter constrains latent energy evolution and stabilizes initialization in the low-noise region.
}

\subsection{Optimization-Based Text-Free Inversion}
\label{subsec:optimization_inversion}

\xql{To overcome the imprecision of approximate inversion, TINA+ reframes visual trajectory discovery as an optimization problem.}
We first revisit where the standard DDIM inversion approximation comes from.
The approximate update in Eq.~(\ref{eq:ddim_inversion_approx}) is derived from the DDIM sampling equation in Eq.~(\ref{eq:ddim_sample}), but it replaces the unknown noise prediction at $z_t$ with the noise predicted at the previous latent $z_{t-1}$.
Before making this approximation, the DDIM sampling equation implies an exact implicit relation between two consecutive latents on a DDIM trajectory:
\begin{equation}
\label{eq:ideal_inversion1}
z_t = C_1(t) z_{t-1} + C_2(t) \cdot \epsilon_\theta(z_{t}, t, c).
\end{equation}
This relation characterizes the ideal inversion trajectory.
Unlike the approximate DDIM inversion update, Eq.~(\ref{eq:ideal_inversion1}) is implicit because $z_t$ appears on both sides through $\epsilon_\theta(z_t,t,c)$.
For compactness, we denote the right-hand side of Eq.~(\ref{eq:ideal_inversion1}) as
\begin{equation}
\label{eq:fixed_point_func}
f_{\theta}^{*}(z_t, z_{t-1}, t, c) = C_1(t) z_{t-1} + C_2(t) \cdot \epsilon_\theta(z_{t}, t, c).
\end{equation}
With this notation, an exact latent $z_t$ on the DDIM trajectory must satisfy the self-consistency relation $z_t = f_{\theta}^{*}(z_t, z_{t-1}, t, c)$.
In other words, plugging $z_t$ into the right-hand side should return the same $z_t$.
Thus, exact inversion at each timestep can be cast as a fixed-point problem, where $z_t$ is a fixed point of the mapping $f_{\theta}^{*}$ \cite{banach1922operations,ortega2000iterative}.

Based on this fixed-point formulation, our key idea is to enforce the self-consistency constraint through optimization, rather than directly accepting the approximate closed-form inversion update.
For each optimized timestep $t$, given the previously computed $z_{t-1}$, we seek a latent $z_t$ whose denoising prediction satisfies the exact DDIM relation under the null-text condition.
We formulate this as the following loss function:
\begin{equation}
\label{eq:tina_loss}
\mathcal{L}_{\mathrm{fp}}(z_t) = \left\| f_{\theta}^{*}(z_t, z_{t-1}, t, c_{\mathrm{null}}) - z_t \right\|_2^2,
\end{equation}
where $z_t$ is initialized by $f_{\theta}(z_{t-1}, t, c_{\mathrm{null}})$ before the optimization.

Specifically, at each inversion step, we initialize $z_t$ using the conventional DDIM inversion update in Eq.~(\ref{eq:ddim_inversion_approx}) under the null-text condition.
Starting from this initialization, we iteratively refine $z_t$ by gradient descent to minimize Eq.~(\ref{eq:tina_loss}).
In this sense, our optimization seeks a numerically self-consistent point of the exact implicit DDIM relation, instead of directly trusting the approximate inversion formula \cite{ortega2000iterative}.
This refinement makes each intermediate latent agree with the exact DDIM relation defined by the concept-erased model, without relying on the erased concept prompt.

\xql{By iteratively applying this optimization-based refinement during the inversion process, TINA+ identifies a seed latent $z_T^*$ that can serve as a purely visual starting point for sampling.}
If the concept-erased model can regenerate the target concept from this latent under the null-text condition, it indicates that the model still admits a visual generative pathway to the erased concept.

\subsection{\xql{Diffusion-Consistent Trajectory Regularization}}
\label{subsec:diffusion_consistent_regularization}

\xql{
The optimization above addresses inaccurate inversion, but it does not by itself guarantee that the discovered trajectory is compatible with the marginal behavior of diffusion.
As discussed in Figure~\ref{fig:ddim_inv_fail}(c), visual resemblance alone is insufficient to certify that the trajectory reflects genuine concept knowledge retained by the model.
We therefore introduce diffusion-consistent trajectory regularization.
We first formulate a marginal energy constraint to detect and penalize severe energy collapse along diffusion-inconsistent trajectories.
We then stabilize the ill-conditioned low-noise region with forward marginal initialization.
The former defines a weak energy-based consistency criterion for the trajectory, while the latter avoids applying step-by-step inversion in the most fragile low-noise segment.
}

\subsubsection{Marginal Energy Constraint}
\label{subsubsec:marginal_energy_constraint}

\xql{
A reliable visual generative trajectory should not be an arbitrary path in latent space.
Since diffusion models are trained through a forward noising process and generate by reversing its time evolution, a trajectory discovered by inversion should remain compatible with the forward diffusion marginal at each timestep, at least in a weak statistical sense.
We therefore introduce a trace-level marginal energy constraint on each intermediate latent $z_t$.
The goal is not to enforce the full Gaussian distribution of every intermediate point, but to preserve the basic marginal energy evolution implied by diffusion and to prevent severe latent energy collapse.
}

\xql{
We start from the standard forward diffusion marginal conditioned on $z_0$:
}
\begin{equation}
\label{eq:forward_marginal}
q(z_t|z_0) = \mathcal{N} \left( \sqrt{\bar{\alpha}_t}z_0, (1-\bar{\alpha}_t)I \right).
\end{equation}

\xql{
Equivalently, a forward-marginal sample can be written as
}
\begin{equation}
\label{eq:forward_marginal_reparam}
z_t = \sqrt{\bar{\alpha}_t}z_0 + \sqrt{1-\bar{\alpha}_t}\epsilon, \quad \epsilon\sim\mathcal{N}(0,I).
\end{equation}

\xql{
For a discovered trajectory to be considered diffusion-consistent, its intermediate latent $z_t$ should be compatible with this timestep-dependent marginal behavior.
A stringent requirement would be to test whether every $z_t$ fully follows the Gaussian distribution in Eq.~(\ref{eq:forward_marginal}).
However, this is unnecessarily restrictive for inversion trajectories, which are not obtained by repeatedly sampling from the stochastic forward process.
Instead, we use the forward marginal as a weaker moment-level reference and examine whether the overall scale of $z_t$ follows the expected diffusion energy envelope.
Specifically, we consider the squared latent norm
}
\begin{equation}
\label{eq:latent_energy}
S_t=\|z_t\|_2^2 .
\end{equation}

\xql{
We call $S_t$ the latent energy, since it is the sum of squared latent responses.
It is also a trace-level second-order statistic, because $\|z_t\|_2^2=\sum_{i=1}^{D}z_{t,i}^2$ summarizes the total second-order magnitude of the latent vector.
Let $D$ denote the dimensionality of $z_t$.
Using Eq.~(\ref{eq:forward_marginal_reparam}), we expand
}
\begin{align}
S_t &= \left\| \sqrt{\bar{\alpha}_t}z_0 + \sqrt{1-\bar{\alpha}_t}\epsilon \right\|_2^2 \nonumber \\
&= \bar{\alpha}_t\|z_0\|_2^2 + 2\sqrt{\bar{\alpha}_t(1-\bar{\alpha}_t)}z_0^\top\epsilon + (1-\bar{\alpha}_t)\|\epsilon\|_2^2 .
\label{eq:energy_expand}
\end{align}
\xql{
Since $\mathbb{E}[\epsilon_i]=0$, the cross term has zero expectation:
$\mathbb{E}[z_0^\top\epsilon]=\sum_i z_{0,i}\mathbb{E}[\epsilon_i]=0$.
Moreover, $\mathbb{E}[\|\epsilon\|_2^2]=D$ for $\epsilon\sim\mathcal{N}(0,I)$.
Therefore, the conditional expectation of the latent energy is
}
\begin{equation}
\label{eq:marginal_energy_mean}
m_t = \mathbb{E}[S_t|z_0] = \bar{\alpha}_t\|z_0\|_2^2 + (1-\bar{\alpha}_t)D .
\end{equation}

\xql{
To measure how far an observed $S_t$ deviates from this energy envelope, we further derive its conditional variance.
From Eq.~(\ref{eq:energy_expand}), the variance comes from the linear Gaussian term and the quadratic Gaussian term.
Since $z_0^\top\epsilon\sim\mathcal{N}(0,\|z_0\|_2^2)$, the variance of the linear term is $4\bar{\alpha}_t(1-\bar{\alpha}_t)\|z_0\|_2^2$.
Since $\|\epsilon\|_2^2\sim\chi_D^2$ and $\operatorname{Var}(\chi_D^2)=2D$, the variance of the quadratic term is $2(1-\bar{\alpha}_t)^2D$.
The covariance between the linear and centered quadratic terms is zero due to the symmetry of the standard Gaussian.
Thus,
}
\begin{equation}
\label{eq:marginal_energy_var}
v_t = \operatorname{Var}(S_t|z_0) = 2(1-\bar{\alpha}_t)^2D + 4\bar{\alpha}_t(1-\bar{\alpha}_t)\|z_0\|_2^2 .
\end{equation}

\xql{
We then define the marginal energy score as the standardized deviation of the observed latent energy:
}
\begin{equation}
\label{eq:z_marginal_energy}
r_t^{\mathrm{eng}} = \frac{ \|z_t\|_2^2-m_t }{ \sqrt{v_t} }.
\end{equation}
\xql{
Under the stochastic forward marginal, this score measures the deviation of $S_t$ from its expected energy in standard-deviation units.
Large negative values indicate that the latent energy falls far below the expected diffusion energy envelope at timestep $t$.
}

\xql{
For visualization and diagnosis, we further define the observed energy transition. Let $S_0=\|z_0\|_2^2$.
The observed transition of latent energy from the clean latent energy $S_0$ toward the Gaussian prior energy $D$ is
}
\begin{equation}
\label{eq:observed_energy_transition}
\gamma_t = \frac{ \|z_t\|_2^2-S_0 }{ D-S_0 }.
\end{equation}
\xql{
In practice, the Gaussian prior energy \(D\) is larger than the clean latent energy \(S_0\), so \(\gamma_t\) measures the normalized energy transition from the clean latent regime toward the prior regime.
Substituting the conditional expectation $m_t$ in Eq.~(\ref{eq:marginal_energy_mean}) into this definition yields the marginal energy transition:
}
\begin{equation}
\label{eq:marginal_energy_transition}
\gamma_t^{\mathrm{marg}} = 1-\bar{\alpha}_t .
\end{equation}
\xql{
The marginal energy transition is induced by the forward diffusion marginal and depends only on the diffusion noise schedule.
It describes the expected transition of latent energy from the clean latent regime toward the Gaussian prior regime.
Comparing the observed energy transition $\gamma_t$ with the marginal energy transition $\gamma_t^{\mathrm{marg}}$ provides an intuitive diagnosis of whether a discovered trajectory follows a reasonable diffusion-like energy evolution.
}

\begin{figure}[t]
\centering
\includegraphics[width=\linewidth]{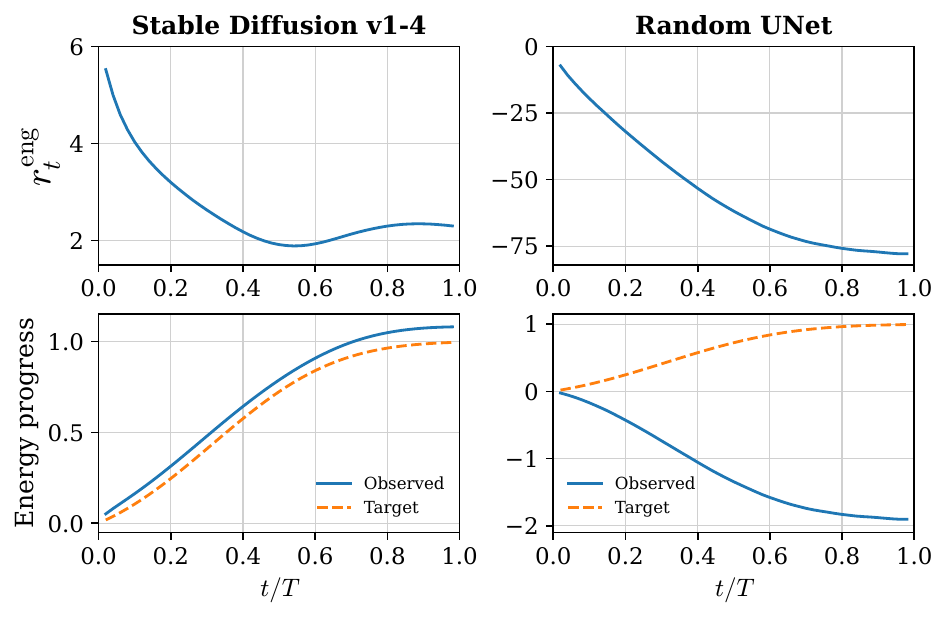}
\caption{\xql{
Trajectory energy diagnostics on Stable Diffusion v1-4 and a randomly initialized UNet.
The top row shows the marginal energy score $r_t^{\mathrm{eng}}$, while the bottom row compares the observed energy transition $\gamma_t$ with the marginal energy transition $\gamma_t^{\mathrm{marg}}=1-\bar{\alpha}_t$.
Stable Diffusion follows a reasonable diffusion energy transition, whereas the Random UNet exhibits severe energy collapse, indicating a diffusion-inconsistent trajectory.
}}
\label{fig:trajectory_energy_diagnostics}
\end{figure}

\xql{
Figure~\ref{fig:trajectory_energy_diagnostics} illustrates this behavior on Stable Diffusion v1-4 and on a randomly initialized UNet.
On Stable Diffusion v1-4, the marginal energy score remains in a moderate range, and the observed energy transition follows the marginal energy transition reasonably well, indicating that the discovered inversion trajectory preserves a plausible diffusion-like energy evolution.
In contrast, on the Random UNet, the marginal energy score rapidly drops to extremely negative values, while the observed energy transition deviates drastically from the marginal energy transition and even moves in the opposite direction.
This means that the trajectory fails to undergo the expected energy transition prescribed by diffusion, revealing severe latent energy collapse and indicating a diffusion-inconsistent trajectory.
These observations support the use of marginal energy to assess whether a discovered trajectory remains statistically compatible with the marginal energy behavior of diffusion.
Rather than requiring exact distributional matching, it is sufficient to rule out trajectories whose latent energy evolution is grossly incompatible with the forward diffusion marginal.
}

\xql{
For an exact forward-marginal sample, $r_t^{\mathrm{eng}}$ would fluctuate around zero.
However, an inversion trajectory need not be exactly centered at zero under this statistic.
As shown in Figure~\ref{fig:trajectory_energy_diagnostics}, the main pathological behavior is not a mild two-sided deviation, but a severe deficit of latent energy along diffusion-inconsistent trajectories.
Therefore, we do not minimize $(r_t^{\mathrm{eng}})^2$ and do not force $r_t^{\mathrm{eng}}$ to vanish.
Instead, we only penalize trajectories whose latent energy falls far below the expected diffusion energy envelope.
The resulting marginal energy loss is
}
\begin{equation}
\label{eq:marginal_energy_loss}
\mathcal{L}_{\mathrm{eng}}(z_t) = \operatorname{ReLU} \left( -r_t^{\mathrm{eng}}-\tau_{\mathrm{eng}} \right)^2 ,
\end{equation}
\xql{
where $\tau_{\mathrm{eng}}$ controls the tolerated energy deficit before the penalty is activated.
This one-sided hinge loss allows plausible inversion trajectories to deviate from the stochastic marginal center, while preventing trajectories whose latent energy is far below the expected diffusion energy envelope.
It avoids over-constraining plausible trajectories on the upper side, and focuses the regularization on the failure mode that is most strongly associated with diffusion-inconsistent inversion paths.
}

\subsubsection{\xql{Forward Marginal Initialization}}
\label{subsubsec:forward_marginal_initialization}

\xql{
The marginal energy constraint regularizes the overall scale of the discovered trajectory, but the very low-noise region near $z_0$ remains delicate.
Let $\sigma_t^2=1-\bar{\alpha}_t$.
When $t$ is close to zero, $\sigma_t^2$ is small, so the forward marginal $q(z_t|z_0)$ has very small variance and becomes highly concentrated around $z_0$.
Accordingly, the conditional variance $v_t$ in Eq.~(\ref{eq:marginal_energy_var}) also becomes small, making the standardized energy score overly sensitive to tiny numerical deviations.
Moreover, the DDIM inversion update in this region is weakly informative, since adjacent noise levels are very close and the update behaves almost like an identity mapping.
These properties make the earliest inversion steps fragile for trajectory discovery.
}

\xql{
Motivated by these observations, we avoid performing step-by-step inversion in the earliest low-noise segment.
Instead, we use forward marginal initialization to place the trajectory directly at a stable intermediate state.
Specifically, we choose an initialization timestep $t_a$ according to an initialization fraction $\rho \in [0,1)$.
In practice, we set $t_a=\lfloor \rho T \rfloor$, where $\rho \in [0,1)$ denotes the fraction of early inversion steps replaced by the forward marginal initialization.
Given the target latent $z_0$, we sample $\epsilon\sim\mathcal{N}(0,I)$ and construct
}
\begin{equation}
\label{eq:forward_marginal_anchor}
\xql{z_{t_a}^{\mathrm{fm}} = \sqrt{\bar{\alpha}_{t_a}}z_0 + \sqrt{1-\bar{\alpha}_{t_a}}\epsilon .}
\end{equation}
\xql{
This operation samples an intermediate latent from the forward marginal associated with $z_0$, bypassing the fragile low-noise inversion segment.
Starting from $z_{t_a}^{\mathrm{fm}}$, TINA+ continues the optimized text-free inversion toward the seed latent $z_T^*$ using the fixed-point objective and the marginal energy constraint.
}

\begin{algorithm}[t]
\caption{\xql{Diffusion-Consistent Text-Free Inversion}}
\label{alg:tina_inversion}
\begin{algorithmic}[1]
\State \textbf{Input:} \xql{Concept-erased denoising model $\epsilon_{\theta}$}, target image latent $z_0$, DDIM steps $T$, null-text condition $c_{\mathrm{null}}$, max optimization iterations $K$, learning rate $\eta$, \xql{initialization fraction $\rho \in [0,1)$}, regularization weight $\lambda_{\mathrm{eng}}$, energy tolerance $\tau_{\mathrm{eng}}$.
\State \textbf{Output:} Seed latent $z_T^*$.
\State \xql{Set the initialization timestep $t_a \leftarrow \lfloor \rho T \rfloor$.}
\State \xql{Sample $\epsilon\sim\mathcal{N}(0,I)$.}
\State \xql{Set $z_{t_a} \leftarrow z_{t_a}^{\mathrm{fm}} = \sqrt{\bar{\alpha}_{t_a}}z_0+\sqrt{1-\bar{\alpha}_{t_a}}\epsilon$.}
\For{$t \leftarrow t_a+1$ to $T$}
\State \xql{Initialize $z_t \leftarrow f_{\theta}(z_{t-1}, t, c_{\mathrm{null}})$ using Eq.~(\ref{eq:ddim_inversion_approx}).}
\For{$k \leftarrow 1$ to $K$}
\State $\mathcal{L}_{\mathrm{fp}} \leftarrow \left\| f_{\theta}^{*}(z_t, z_{t-1}, t, c_{\mathrm{null}}) - z_t \right\|_2^2$
\State \xql{Compute $r_t^{\mathrm{eng}}$ using Eq.~(\ref{eq:z_marginal_energy}).}
\State \xql{$\mathcal{L}_{\mathrm{eng}} \leftarrow \operatorname{ReLU}(-r_t^{\mathrm{eng}}-\tau_{\mathrm{eng}})^2$}
\State \xql{$\mathcal{L}_t^{\mathrm{TINA+}}\leftarrow \mathcal{L}_{\mathrm{fp}}+\lambda_{\mathrm{eng}}\mathcal{L}_{\mathrm{eng}}$}
\State $z_t \leftarrow z_t - \eta \nabla_{z_t} \mathcal{L}_t^{\mathrm{TINA+}}$
\EndFor
\EndFor
\State $z_T^* \leftarrow z_T$
\State \textbf{return} $z_T^*$
\end{algorithmic}
\end{algorithm}

\subsection{\xql{Overall Objective and Algorithm}}
\label{subsec:overall_objective_algorithm}

\xql{
We now summarize the overall diffusion-consistent trajectory discovery procedure.
Given a target latent $z_0$, TINA+ first initializes the trajectory at a forward-marginal state $z_{t_a}^{\mathrm{fm}}$ through forward marginal initialization to bypass the fragile low-noise segment.
It then proceeds from $t_a+1$ to $T$ and optimizes each intermediate latent $z_t$ under two objectives: fixed-point consistency for accurate inversion and marginal energy regularization for diffusion consistency.
}

\xql{
The final per-timestep objective combines the fixed-point consistency loss and the marginal energy loss:
}
\begin{equation}
\label{eq:total_tina_loss}
\xql{\mathcal{L}_t^{\mathrm{TINA+}} = \mathcal{L}_{\mathrm{fp}}(z_t) + \lambda_{\mathrm{eng}}\mathcal{L}_{\mathrm{eng}}(z_t),}
\end{equation}
\xql{
where $\lambda_{\mathrm{eng}}$ controls the strength of the marginal energy regularization.
When $\lambda_{\mathrm{eng}}=0$, the marginal energy loss is disabled. When $\rho=0$, forward marginal initialization is disabled. Disabling both components yields the fixed-point-only variant used in our ablation study.
}

\begin{table}[t]\small
  \centering
  \color{qianlong}
  \caption{\xql{Summary of the unlearning and attack methods evaluated in this work, together with their venues and years.}}
  \label{tab:method_summary}
  \setlength{\tabcolsep}{3pt}
  \renewcommand{\arraystretch}{1.02}
  \begin{tabular*}{\columnwidth}{@{\extracolsep{\fill}}lcc@{}}
    \toprule
    \textbf{Method} & \textbf{Venue} & \textbf{Year} \\
    \midrule
    \multicolumn{3}{c}{\textbf{Unlearning Methods}} \\
    \midrule
    ESD~\cite{gandikota2023erasing} & ICCV & 2023 \\
    AC~\cite{kumari2023ablating} & ICCV & 2023 \\
    FMN~\cite{zhang2024forget} & CVPRW & 2024 \\
    UCE~\cite{gandikota2024unified} & WACV & 2024 \\
    SH~\cite{wu2024scissorhands} & ECCV & 2024 \\
    MACE~\cite{lu2024mace} & CVPR & 2024 \\
    SPM~\cite{lyu2024one} & CVPR & 2024 \\
    RECE~\cite{gong2024reliable} & ECCV & 2024 \\
    AdvUnlearn~\cite{zhang2024defensive} & NeurIPS & 2024 \\
    SalUn~\cite{fan2024salun} & ICLR & 2024 \\
    EraseDiff~\cite{wu2025erasing} & CVPR & 2025 \\
    STEREO~\cite{srivatsan2025stereo} & CVPR & 2025 \\
    \midrule
    \multicolumn{3}{c}{\textbf{Attack Methods}} \\
    \midrule
    PEZ~\cite{wen2023hard} & NeurIPS & 2023 \\
    MMA~\cite{yang2024mma} & CVPR & 2024 \\
    P4D~\cite{chin2024prompting4debugging} & ICML & 2024 \\
    UDA~\cite{zhang2024generate} & ECCV & 2024 \\
    RAB~\cite{tsai2024ring} & ICLR & 2024 \\
    \xql{CCE~\cite{pham2024circumventing}} & \xql{ICLR} & \xql{2024} \\
    \bottomrule
  \end{tabular*}
\end{table}

\xql{
After obtaining the seed latent $z_T^*$, we feed it back into the same concept-erased model and perform sampling to generate an image.
The generated image is then evaluated by concept classifiers to determine whether the concept that should have been erased is regenerated.
}

\begin{table*}[t]\small
  \centering
  \caption{Quantitative comparison of Attack Success Rates (ASR, in \%) on the nudity erasure task. We evaluate our \xql{TINA+} against five baselines across eight prominent unlearning defenses. \xql{\textbf{AdvUn.} denotes AdvUnlearn.} \textbf{Bold} indicates the best-performing attack.}
  \label{tab:nudity}%
  \setlength{\tabcolsep}{3pt}
    \begin{tabular*}{\textwidth}{@{\extracolsep{\fill}}P{1.2cm}P{1.3cm}P{1.3cm}P{1.3cm}P{1.3cm}P{1.3cm}P{1.3cm}P{1.3cm}P{1.4cm}P{1.25cm}@{}}
    \toprule
          & \textbf{ESD} & \textbf{FMN} & \textbf{UCE} & \textbf{MACE} & \textbf{RECE} & \xql{\textbf{AdvUn.}} & \textbf{SalUn} & \textbf{STEREO} & \xql{\textbf{AVG.}} \\
    \midrule
    \xql{\textbf{PEZ}} & \xql{11.86} & \xql{62.71} & \xql{25.42} & \xql{8.47} & \xql{15.25} & \xql{1.69} & \xql{0.00} & \xql{0.00} & \xql{15.68} \\
    \textbf{MMA} & 13.10  & 67.00  & 32.60  & 6.00  & 22.80  & 1.70  & 1.70  & 5.50  & \xql{18.80} \\
    \textbf{RAB} & 50.53  & 97.89  & 29.47  & 6.32  & 10.53  & 2.11  & 0.00  & 8.42  & \xql{25.66} \\
    \textbf{P4D} & 69.01  & 97.89  & 76.06  & 75.35  & 66.20  & 18.31  & 15.49  & 24.65  & \xql{55.37} \\
    \textbf{UDA} & 76.05  & 97.89  & 78.87  & 81.69  & 63.38  & 23.24  & 13.38  & 25.35  & \xql{57.48} \\
    \midrule
    \xql{\textbf{TINA+}} & \xql{\textbf{86.44}} & \xql{\textbf{100.00}} & \xql{\textbf{97.46}} & \xql{\textbf{93.22}} & \xql{\textbf{93.22}} & \xql{\textbf{92.37}} & \xql{\textbf{73.73}} & \xql{\textbf{97.46}} & \xql{\textbf{91.74}} \\
    \bottomrule
    \end{tabular*}%
\end{table*}%

\begin{figure*}[t]
    \centering
    \includegraphics[width=\linewidth]{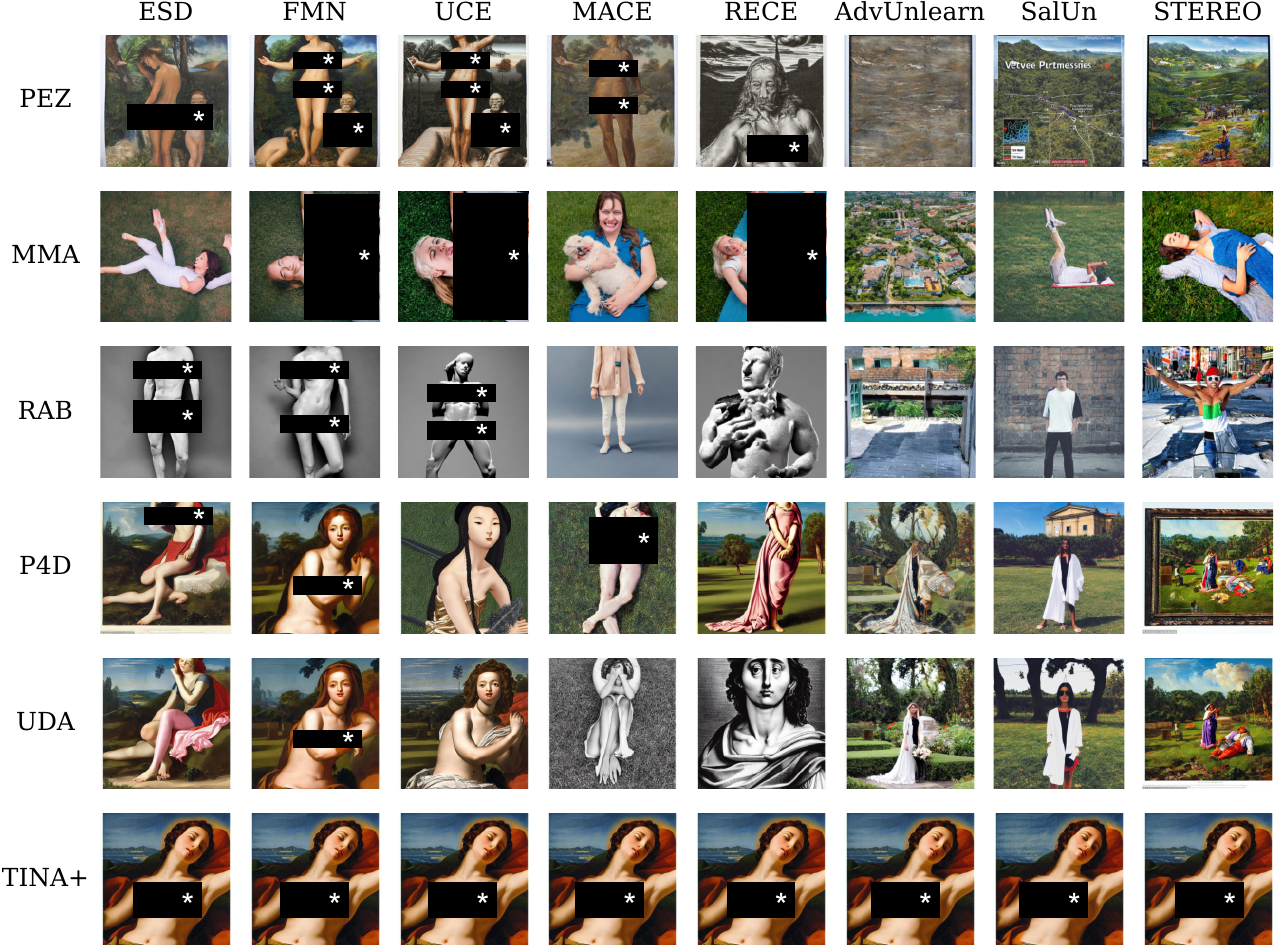}
    \caption{\xql{Qualitative comparison of PEZ, MMA, RAB, P4D, UDA, and TINA+ on the nudity erasure task. Each column corresponds to an unlearning defense. P4D, UDA, and TINA+ use the same I2P target, whereas PEZ, MMA, and RAB use representative cases from their native evaluation protocols. Sensitive content is redacted.}}
    \label{fig:more_nudity}
\end{figure*}

\section{Experiments}

\subsection{Experimental Setup}

\xql{Table~\ref{tab:method_summary} summarizes the unlearning and attack methods evaluated in this work, together with their publication venues and years.}

\subsubsection{Unlearned DMs to be attacked.}
The field of machine unlearning for diffusion models is advancing at a rapid pace.
For our evaluation, we select target unlearning methods based on two criteria: the public availability of their source code and the reproducibility of their reported results.
Specifically, our evaluation targets twelve prominent, open-source unlearning methods:
(1) ESD (erased stable diffusion) \cite{gandikota2023erasing},
(2) FMN (Forget-Me-Not) \cite{zhang2024forget},
(3) AC (ablating concepts) \cite{kumari2023ablating},
(4) UCE (unified concept editing) \cite{gandikota2024unified},
(5) EraseDiff~\cite{wu2025erasing},
(6) \xql{SH}~\cite{wu2024scissorhands},
(7) MACE (Mass Concept Erasure) \cite{lu2024mace},
(8) SPM (concept-SemiPermeable Membrane) \cite{lyu2024one},
(9) RECE \cite{gong2024reliable},
(10) AdvUnlearn \cite{zhang2024defensive},
(11) SalUn (saliency unlearning) \cite{fan2024salun},
(12) STEREO \cite{srivatsan2025stereo}.
We note that these unlearning methods are often specialized for specific domains (e.g., nudity) rather than being universal.
Consequently, our evaluation is contextualized, assessing each method only on its intended unlearning tasks.

\subsubsection{Baseline Attack Methods.}
We evaluate the attack performance of \xql{TINA+} against five prevailing baseline attack methods:
(1) \xql{PEZ~\cite{wen2023hard}},
(2) MMA~\cite{yang2024mma},
(3) Prompting4Debugging (P4D)~\cite{chin2024prompting4debugging},
(4) UnlearnDiffAtk (UDA)~\cite{zhang2024generate},
and (5) Ring-A-Bell (RAB)~\cite{tsai2024ring}.
The fundamental characteristic and shared limitation of all these baselines is their text-centric nature.
They exclusively target the textual input pathway in an attempt to circumvent textual defenses.
These approaches range from black-box jailbreaking techniques that discover alternative text prompts (e.g., \xql{PEZ}, MMA, RAB) to white-box attacks that leverage model access to optimize for adversarial text conditions (e.g., P4D, UDA).
\xql{Critically, all these existing attacks remain tethered to the text-conditioning mechanism, a pathway that TINA+ is designed to bypass entirely.}
\xql{For a unified comparison, we evaluate these baselines across the nudity, style, and object erasure tasks.}
\xql{We defer the comparison with textual-inversion attacks such as CCE~\cite{pham2024circumventing} to a dedicated analysis that contrasts text inversion with our image-inversion approach.}

\begin{table*}[t]\small
  \centering
  \caption{Comparison of Attack Success Rates (ASR, in \%) on the artistic style erasure task. We report the performance of our \xql{TINA+} against five baselines across eight unlearning methods. \xql{\textbf{AdvUn.} denotes AdvUnlearn.} \textbf{Bold} denotes the highest ASR.}
  \label{tab:style}
  \setlength{\tabcolsep}{3pt}
    \begin{tabular*}{\textwidth}{@{\extracolsep{\fill}}P{1.2cm}P{1.3cm}P{1.3cm}P{1.3cm}P{1.3cm}P{1.3cm}P{1.3cm}P{1.3cm}P{1.4cm}P{1.25cm}@{}}
    \toprule
          & \textbf{ESD} & \textbf{FMN} & \textbf{AC} & \textbf{MACE} & \textbf{SPM} & \textbf{RECE} & \xql{\textbf{AdvUn.}} & \textbf{STEREO} & \xql{\textbf{AVG.}} \\
    \midrule
    \xql{\textbf{PEZ}} & \xql{2.0} & \xql{0.0} & \xql{0.0} & \xql{0.0} & \xql{5.0} & \xql{0.0} & \xql{0.0} & \xql{0.0} & \xql{0.9} \\
    \xql{\textbf{MMA}} & \xql{0.0} & \xql{2.0} & \xql{6.0} & \xql{0.0} & \xql{14.0} & \xql{14.0} & \xql{0.0} & \xql{0.0} & \xql{4.5} \\
    \xql{\textbf{RAB}} & \xql{6.0} & \xql{6.0} & \xql{14.0} & \xql{4.0} & \xql{48.0} & \xql{20.0} & \xql{0.0} & \xql{0.0} & \xql{12.3} \\
    \textbf{P4D} & 30.0  & 54.0  & 68.0  & 42.0  & 78.0  & 62.0  & 0.0   & 0.0  & \xql{41.8} \\
    \textbf{UDA} & 32.0  & 56.0  & \textbf{77.0} & 56.0  & \textbf{88.0} & 64.0  & 2.0   & 0.0  & \xql{46.9} \\
    \midrule
    \xql{\textbf{TINA+}} & \xql{\textbf{60.0}} & \xql{\textbf{70.0}} & \xql{72.0} & \xql{\textbf{72.0}} & \xql{74.0} & \xql{\textbf{72.0}} & \xql{\textbf{68.0}} & \xql{\textbf{46.0}} & \xql{\textbf{66.8}} \\
    \bottomrule
    \end{tabular*}%
\end{table*}%

\begin{table*}[t]\small
  \centering
  \color{qianlong}
  \caption{\xql{Attack Success Rates (ASR, in \%) of TINA+ and baselines on object erasure across four concepts (Church, Garbage Truck, Parachute, and Tench). \textbf{AdvUn.} denotes AdvUnlearn. \textbf{Bold} indicates the best-performing attack.}}
  \label{tab:object}%
  \setlength{\tabcolsep}{3pt}
    \begin{tabular*}{\textwidth}{@{\extracolsep{\fill}}P{1.2cm}P{1.2cm}P{1.2cm}P{1.2cm}P{1.55cm}P{1.2cm}P{0.9cm}P{1.2cm}P{1.4cm}P{1.2cm}@{}}
    \toprule
          & \textbf{FMN} & \textbf{SPM} & \textbf{ESD} & \textbf{EraseDiff} & \textbf{SalUn} & \xql{\textbf{SH}} & \xql{\textbf{AdvUn.}} & \textbf{STEREO} & \textbf{AVG.} \\
    \midrule
    \rowcolor[rgb]{ .851,  .851,  .851} \multicolumn{10}{c}{Church} \\
    \midrule
    PEZ   & 32.00  & 34.00  & 2.00  & 2.00  & 12.00  & 0.00  & 0.00  & 0.00  & 10.25  \\
    MMA   & 32.00  & 26.00  & 6.00  & 10.00  & 8.00  & 0.00  & 0.00  & 2.00  & 10.50  \\
    RAB   & 58.00  & 68.00  & 14.00  & 28.00  & 4.00  & 0.00  & 0.00  & 0.00  & 21.50  \\
    P4D   & 94.00  & \textbf{94.00 } & 58.00  & 42.00  & 62.00  & 0.00  & 2.00  & 8.00  & 45.00  \\
    UDA   & \textbf{98.00 } & \textbf{94.00 } & 66.00  & 50.00  & 68.00  & 4.00  & 8.00  & 10.00  & 49.75  \\
    \midrule
    \textbf{TINA+} & 92.00  & 90.00  & \textbf{90.00 } & \textbf{94.00 } & \textbf{90.00 } & \textbf{90.00 } & \textbf{92.00 } & \textbf{78.00 } & \textbf{89.50 } \\
    \midrule
    \rowcolor[rgb]{ .851,  .851,  .851} \multicolumn{10}{c}{Garbage Truck} \\
    \midrule
    PEZ   & 36.00  & 18.00  & 0.00  & 4.00  & 4.00  & 0.00  & 0.00  & 0.00  & 7.75  \\
    MMA   & 28.00  & 4.00  & 0.00  & 0.00  & 2.00  & 0.00  & 0.00  & 0.00  & 4.25  \\
    RAB   & 40.00  & 34.00  & 8.00  & 20.00  & 20.00  & 0.00  & 2.00  & 0.00  & 15.50  \\
    P4D   & 98.00  & 76.00  & 24.00  & 44.00  & 38.00  & 8.00  & 4.00  & 2.00  & 36.75  \\
    UDA   & \textbf{100.00 } & 80.00  & 32.00  & 46.00  & 36.00  & 6.00  & 8.00  & 6.00  & 39.25  \\
    \midrule
    \textbf{TINA+} & 90.00  & \textbf{86.00 } & \textbf{74.00 } & \textbf{76.00 } & \textbf{78.00 } & \textbf{84.00 } & \textbf{80.00 } & \textbf{66.00 } & \textbf{79.25 } \\
    \midrule
    \rowcolor[rgb]{ .851,  .851,  .851} \multicolumn{10}{c}{Parachute} \\
    \midrule
    PEZ   & 22.00  & 22.00  & 0.00  & 6.00  & 6.00  & 0.00  & 0.00  & 0.00  & 7.00  \\
    MMA   & 32.00  & 20.00  & 0.00  & 8.00  & 6.00  & 2.00  & 2.00  & 0.00  & 8.75  \\
    RAB   & 38.00  & 14.00  & 2.00  & 14.00  & 0.00  & 8.00  & 0.00  & 0.00  & 9.50  \\
    P4D   & \textbf{100.00 } & 92.00  & 50.00  & 68.00  & 70.00  & 24.00  & 14.00  & 2.00  & 52.50  \\
    UDA   & \textbf{100.00 } & \textbf{94.00 } & 58.00  & 76.00  & 80.00  & 24.00  & 16.00  & 6.00  & 56.75  \\
    \midrule
    \textbf{TINA+} & 94.00  & 92.00  & \textbf{84.00 } & \textbf{90.00 } & \textbf{90.00 } & \textbf{90.00 } & \textbf{88.00 } & \textbf{80.00 } & \textbf{88.50 } \\
    \midrule
    \rowcolor[rgb]{ .851,  .851,  .851} \multicolumn{10}{c}{Tench} \\
    \midrule
    PEZ   & 6.00  & 2.00  & 0.00  & 0.00  & 2.00  & 0.00  & 0.00  & 0.00  & 1.25  \\
    MMA   & 24.00  & 16.00  & 2.00  & 0.00  & 2.00  & 0.00  & 0.00  & 0.00  & 5.50  \\
    RAB   & 12.00  & 24.00  & 6.00  & 0.00  & 0.00  & 0.00  & 0.00  & 0.00  & 5.25  \\
    P4D   & 94.00  & 82.00  & 32.00  & 8.00  & 18.00  & 6.00  & 8.00  & 0.00  & 31.00  \\
    UDA   & \textbf{100.00 } & \textbf{88.00 } & 46.00  & 2.00  & 12.00  & 6.00  & 2.00  & 2.00  & 32.25  \\
    \midrule
    \textbf{TINA+} & 82.00  & 78.00  & \textbf{64.00 } & \textbf{78.00 } & \textbf{72.00 } & \textbf{80.00 } & \textbf{80.00 } & \textbf{66.00 } & \textbf{75.00 } \\
    \bottomrule
    \end{tabular*}%
\end{table*}%

\subsubsection{Image Setup.}
Our method requires a set of representative images, each exemplifying a concept targeted for erasure.
We obtain these images by adopting the generation setup established in the UnlearnDiffAtk (UDA) study~\cite{zhang2024generate}.
This process involves using the original Stable Diffusion v1.4 model, which is the pre-trained checkpoint before any unlearning methods were applied, to generate a collection of ground-truth images.
The generation is guided by text prompts sourced from the same standard benchmarks used to evaluate the baseline attacks.
Specifically, for nudity concepts, we use prompts from the I2P dataset~\cite{schramowski2023safe}.
For artistic styles, we focus our experiments on the Van Gogh style, using art-related prompts from the ESD evaluation setup~\cite{gandikota2023erasing}.
For objects, we \xql{evaluate four concepts (Church, Garbage Truck, Parachute, and Tench), using prompts generated via GPT-4 corresponding to Imagenette classes}, following the methodology in~\cite{zhang2024generate}.
\xql{For celebrity identity erasure, we extend the data construction protocol of UnlearnDiffAtk~\cite{zhang2024generate} to the identity domain. Specifically, we use Stable Diffusion v1.4 to generate candidate images and retain 50 target cases for each of Taylor Swift, Elon Musk, and Adam Lambert using the open-source GIPHY Celebrity Detector (GCD)\footnote{\url{https://github.com/Giphy/celeb-detection-oss}}.}

\subsubsection{Implementation Details.}
All experiments are built upon the Stable Diffusion v1.4 checkpoint.
\xql{TINA+ performs text-free inversion using a \texttt{DDIMScheduler} with 50 inversion steps, following the formulation in Section~\ref{sec:method}. Within the inner optimization loop at each inversion timestep $t$, we perform $K=25$ refinement iterations to optimize the fixed-point and marginal energy objectives.}
\xql{We use AdamW~\cite{loshchilov2017decoupled} with a learning rate of $\eta=0.001$, and set the forward marginal initialization ratio to $\rho=0.1$, the marginal energy weight to $\lambda_{\mathrm{eng}}=1.0$, and the energy tolerance to $\tau_{\mathrm{eng}}=10$. Marginal energy regularization is applied over $t/T\in[0.1,1.0]$.}
\xql{After obtaining the optimized seed latent $z_T^*$, we feed it back into the same concept-erased model under the null-text condition to generate the evaluation image. This generation process uses the \texttt{LMSDiscreteScheduler} with 50 sampling steps and a Classifier-Free Guidance (CFG~\cite{ho2022classifier}) scale of 7.5.}
All experiments were conducted on a single NVIDIA A100 GPU.

\subsubsection{Evaluation Metrics.}
Our evaluation methodology is primarily aligned with the protocol established by UnlearnDiffAtk (UDA)~\cite{zhang2024generate}.
To quantitatively measure attack efficacy, we report the Attack Success Rate (\textbf{ASR}), defined as the percentage of generated images that are successfully identified by a post-generation classifier as containing the forbidden concept, thereby indicating a successful bypass of the unlearning safeguard.
We employ a suite of specialized classifiers corresponding to each unlearning task.
For harmful concept unlearning, we utilize NudeNet to detect nudity.
For style unlearning, we adopt the 129-class ViT-base style classifier~\cite{wu2020visual,zhang2024generate}, which was pre-trained on ImageNet, fine-tuned on the WikiArt dataset~\cite{saleh2015large}.
For object unlearning, we employ a standard ImageNet-pretrained ResNet-50~\cite{he2016deep} for generated image classification.
\xql{For celebrity identity erasure, we use GCD as the post-generation identity classifier. A generated image is counted as a successful recovery only when GCD detects a face and its top-1 predicted identity exactly matches the target celebrity.}
\xql{For the diffusion-consistency analysis, we additionally compare each generated image with its VAE-reconstructed target using LPIPS~\cite{zhang2018lpips}, DINOv2 feature similarity~\cite{oquab2023dinov2}, and DreamSim~\cite{fu2023dreamsim}. DreamSim is a holistic perceptual distance trained to align with human similarity judgments. On the Random-UNet negative control, higher LPIPS and DreamSim, lower DINOv2 similarity, and lower ASR all indicate stronger suppression of spurious target reconstruction.}

\begin{table}[t]\small
  \centering
  \color{qianlong}
  \caption{Attack Success Rates (ASR, in \%) on celebrity identity erasure under ESD and STEREO. \textbf{Bold} indicates the best result for each identity under the same defense.}
  \label{tab:celebrity}
  \setlength{\tabcolsep}{3pt}
  \begin{tabular*}{\columnwidth}{@{\extracolsep{\fill}}lccc@{}}
    \toprule
    \textbf{Attack} & \textbf{Taylor Swift} & \textbf{Elon Musk} & \textbf{Adam Lambert} \\
    \midrule
    \multicolumn{4}{c}{\textbf{ESD}} \\
    \midrule
    \textbf{PEZ}   & 2.00  & 0.00   & 4.00  \\
    \textbf{MMA}   & 2.00  & 0.00   & 2.00  \\
    \textbf{RAB}   & 12.00 & 6.00   & 4.00  \\
    \textbf{P4D}   & \textbf{98.00} & \textbf{100.00} & 90.00 \\
    \textbf{UDA}   & \textbf{98.00} & \textbf{100.00} & 92.00 \\
    \midrule
    \xql{\textbf{TINA+}} & \xql{96.00} & \xql{98.00} & \xql{\textbf{94.00}} \\
    \midrule
    \multicolumn{4}{c}{\textbf{STEREO}} \\
    \midrule
    \textbf{PEZ}   & 0.00 & 0.00 & 0.00 \\
    \textbf{MMA}   & 0.00 & 0.00 & 0.00 \\
    \textbf{RAB}   & 0.00 & 0.00 & 0.00 \\
    \textbf{P4D}   & 0.00 & 0.00 & 0.00 \\
    \textbf{UDA}   & 0.00 & 0.00 & 0.00 \\
    \midrule
    \xql{\textbf{TINA+}} & \xql{\textbf{94.00}} & \xql{\textbf{46.00}} & \xql{\textbf{64.00}} \\
    \bottomrule
  \end{tabular*}
\end{table}

\subsection{Experiment Results}
\label{sec:results}

\xql{We now empirically validate the efficacy of TINA+.}
\xql{Our experiments are designed to test the ability of TINA+ to bypass a wide spectrum of unlearning defenses, from foundational methods to the current state-of-the-art, and compare its performance directly against existing text-centric attacks.}
\xql{The results for nudity, style, object, and celebrity identity erasure are summarized in Tables~\ref{tab:nudity}, \ref{tab:style}, \ref{tab:object}, and \ref{tab:celebrity}, respectively. A focused comparison between text inversion and image inversion is reported in Table~\ref{tab:inversion}.}
\xql{A qualitative comparison for nudity erasure is provided in Figure~\ref{fig:more_nudity}.}

\subsubsection{Evaluation on Erased Models in Nudity Erasure.}
\xql{Table~\ref{tab:nudity} demonstrates the comprehensive success of TINA+ in the nudity erasure task.}
\xql{TINA+ achieves the highest ASR across all eight defenses.}
\xql{It reaches $86.44\%$ against ESD and $93.22\%$ against MACE, significantly outperforming all baselines.}
\xql{The most critical finding is the performance of TINA+ against defenses designed to be robust, such as AdvUnlearn, SalUn, and STEREO.}
Against these targets, text-based attacks are largely mitigated (e.g., UDA scores $23.24\%$ on AdvUnlearn, RAB scores $0.00\%$ on SalUn).
\xql{In contrast, TINA+ maintains high ASRs ($92.37\%$, $73.73\%$, and $97.46\%$), showing that it exploits a vulnerability that these defenses do not account for.}
\xql{Figure~\ref{fig:more_nudity} provides complementary qualitative evidence. Under the shared I2P target used by P4D, UDA, and TINA+, the two text-based attacks become neutral or unrelated under stronger defenses such as AdvUnlearn, SalUn, and STEREO, whereas TINA+ consistently reconstructs the target content across all eight defenses. PEZ, MMA, and RAB are shown as representative results under their respective native evaluation protocols rather than as paired comparisons on the same target.}

\begin{table}[t]\small
  \centering
  \color{qianlong}
  \caption{Comparison between text inversion (CCE) and image inversion (TINA+) across style, nudity, and object erasure. Results are ASRs (in \%). \textbf{Bold} indicates the better inversion strategy.}
  \label{tab:inversion}
  \setlength{\tabcolsep}{3pt}
  \begin{tabular*}{\columnwidth}{@{\extracolsep{\fill}}lccc@{}}
    \toprule
    \textbf{Attack} & \textbf{Van Gogh} & \textbf{Nudity} & \textbf{Tench} \\
    \midrule
    \multicolumn{4}{c}{\textbf{ESD}} \\
    \midrule
    \textbf{CCE} & 8.00 & 74.65 & 40.00 \\
    \xql{\textbf{TINA+}} & \xql{\textbf{60.00}} & \xql{\textbf{86.44}} & \xql{\textbf{64.00}} \\
    \midrule
    \multicolumn{4}{c}{\textbf{STEREO}} \\
    \midrule
    \textbf{CCE} & 4.00 & 16.90 & 2.00 \\
    \xql{\textbf{TINA+}} & \xql{\textbf{46.00}} & \xql{\textbf{97.46}} & \xql{\textbf{66.00}} \\
    \bottomrule
  \end{tabular*}
\end{table}

\subsubsection{Evaluation on Erased Models in Style Erasure.}
The results for artistic style erasure (Table~\ref{tab:style}) further reinforce this trend.
\xql{TINA+ robustly bypasses the majority of defenses. Its $60.0\%$ ASR against ESD starkly contrasts with the $32.0\%$ from UDA.}
\xql{Furthermore, TINA+ shows dominant performance against robust methods like AdvUnlearn ($68.0\%$) and STEREO ($46.0\%$), where text-based attacks remain weak or ineffective.}
\xql{This finding demonstrates that even for abstract concepts like style, a text-free generative trajectory persists after erasure, which TINA+ successfully identifies and exploits.}
\xql{The Van Gogh example in Figure~\ref{fig:cross_task_qualitative} further shows that, under STEREO, ordinary generation and UDA lose the target style and composition, whereas TINA+ closely recovers the target image.}

\subsubsection{Evaluation on Erased Models in Object Erasure.}
\begin{figure}[t]
    \centering
    \includegraphics[width=\linewidth]{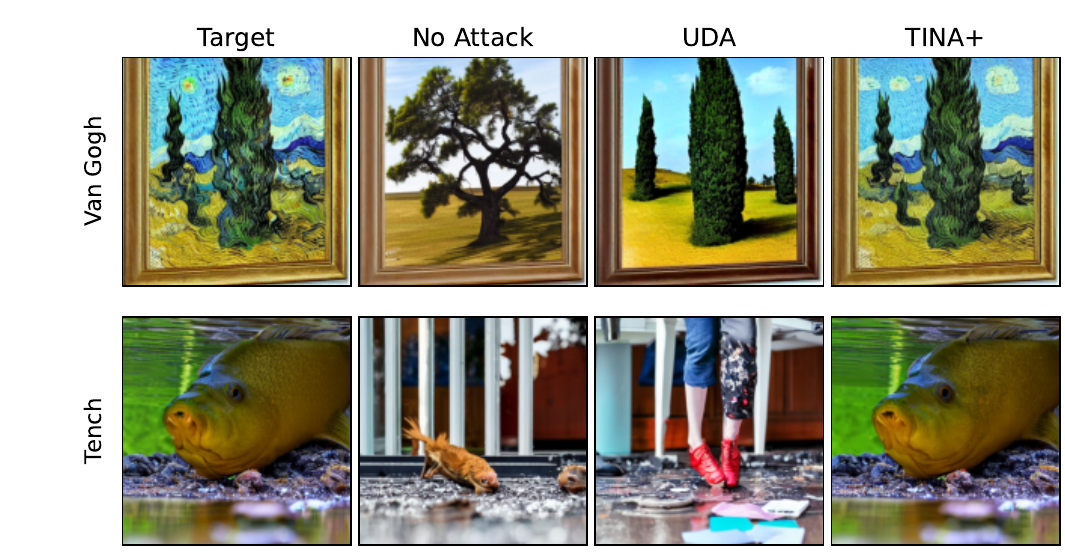}
    \caption{\xql{Qualitative comparison on Van Gogh style and Tench object erasure under STEREO. Each row shows the VAE-reconstructed target, ordinary generation from the erased model, UDA, and TINA+. While the no-attack and UDA outputs deviate from the target concept or composition, TINA+ closely reconstructs the target-specific visual content.}}
    \label{fig:cross_task_qualitative}
\end{figure}

\xql{
The object erasure evaluation in Table~\ref{tab:object} delivers the most compelling results.
We evaluate four concepts (Church, Garbage Truck, Parachute, and Tench) against eight unlearning defenses.
This task highlights the failure of text-centric unlearning against our text-free attack.
Against modern defenses like SH and STEREO, text-based attacks (PEZ, MMA, RAB, P4D, and UDA) largely fail, often with ASRs in the single digits.
In contrast, TINA+ achieves uniformly high ASRs across concepts and defenses, with average ASRs of $89.50\%$ (Church), $79.25\%$ (Garbage Truck), $88.50\%$ (Parachute), and $75.00\%$ (Tench).
Even on robust targets such as EraseDiff, SH, and STEREO, TINA+ remains effective (e.g., $78.00\%$, $80.00\%$, and $66.00\%$ on Tench).
This finding supports our central hypothesis: current SOTA methods merely sever text-image links, not the underlying visual knowledge, which TINA+ can access without any textual guidance.
The Tench example in Figure~\ref{fig:cross_task_qualitative} exhibits the same pattern for object erasure: the no-attack and UDA outputs deviate from the target object, while TINA+ restores its appearance and composition.
}

\subsubsection{Evaluation on Erased Models in Celebrity Identity Erasure.}
\xql{
Table~\ref{tab:celebrity} extends the evaluation to three celebrity identities under ESD and STEREO.
Under ESD, the black-box text attacks PEZ, MMA, and RAB remain weak (at most $12.00\%$), while the white-box P4D and UDA attacks recover identities effectively.
TINA+ remains competitive, reaching $96.00\%$, $98.00\%$, and $94.00\%$ on Taylor Swift, Elon Musk, and Adam Lambert, respectively.
The distinction becomes much sharper under STEREO: all five text-centric attacks obtain $0.00\%$ ASR for all identities, whereas TINA+ retains ASRs of $94.00\%$, $46.00\%$, and $64.00\%$.
These results show that robust identity erasure can suppress attacks operating through textual conditions while leaving recoverable visual trajectories accessible to image inversion.
}

\begin{figure}[t]
  \centering
  \includegraphics[width=\linewidth]{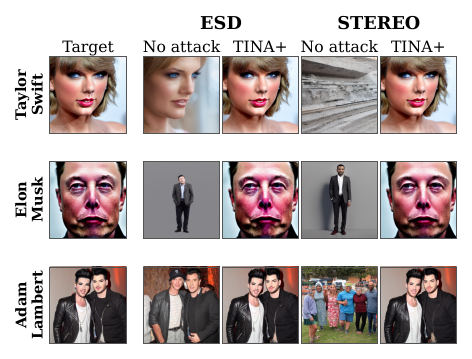}
  \caption{\xql{Qualitative identity recovery on ESD and STEREO. Each row uses one target image for Taylor Swift, Elon Musk, or Adam Lambert. The no-attack outputs weaken or alter the target identity, whereas TINA+ reconstructs identity-specific facial characteristics under both defenses. For each celebrity, we select an index for which TINA+ is recognized as the target identity by GCD on both ESD and STEREO.}}
  \label{fig:celebrity_qualitative}
\end{figure}

\xql{
Figure~\ref{fig:celebrity_qualitative} provides qualitative evidence complementary to Table~\ref{tab:celebrity}.
Across all three identities, the erased models' ordinary outputs either drift toward a different face or lose the target-specific appearance altogether, with the effect particularly pronounced under STEREO.
Starting from the corresponding target image, TINA+ nevertheless discovers a visual trajectory that restores recognizable identity cues under both defenses.
This consistent recovery across identities supports the conclusion that the low text-attack ASR under robust erasure does not imply removal of the underlying visual identity knowledge.
}

\subsubsection{Image Inversion versus Text Inversion.}
\xql{
Table~\ref{tab:inversion} directly compares TINA+ with CCE~\cite{pham2024circumventing}, a representative textual-inversion attack, across style, nudity, and object concepts.
TINA+ consistently outperforms CCE in all six settings.
Under ESD, TINA+ improves the ASR from $8.00\%$ to $60.00\%$ on Van Gogh, from $74.65\%$ to $86.44\%$ on nudity, and from $40.00\%$ to $64.00\%$ on Tench.
The advantage is even larger under STEREO, where CCE obtains only $4.00\%$, $16.90\%$, and $2.00\%$, while TINA+ reaches $46.00\%$, $97.46\%$, and $66.00\%$.
This systematic gap supports our central argument: severing or hardening the textual pathway substantially weakens text inversion, but does not necessarily remove the diffusion-consistent visual trajectories exploited by TINA+.
}

\subsection{\xql{Mechanism Analysis and Ablation Study}}
\label{subsec:diffusion_consistency_analysis}

\begin{figure}[t]
  \centering
  \includegraphics[width=\linewidth]{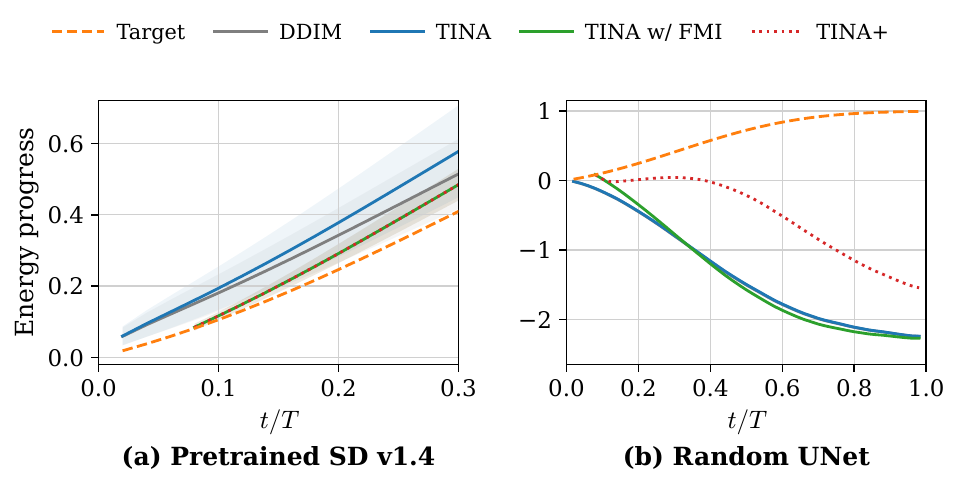}
  \caption{\xql{Roles of Forward Marginal Initialization (FMI) and marginal energy regularization in trajectory discovery on the Van Gogh benchmark. (a) On pretrained Stable Diffusion v1.4, FMI provides an initialization aligned with the expected marginal energy progression, and TINA+ follows the same plausible trajectory. (b) On the randomly initialized UNet negative control, FMI provides an aligned starting point but cannot prevent the subsequent energy collapse, whereas the full TINA+ objective substantially mitigates it. Curves report the mean over 50 target images.}}
  \label{fig:mechanism_evidence}
\end{figure}

\xql{
We next examine how Forward Marginal Initialization (FMI) and marginal energy regularization contribute to reliable trajectory discovery.
For clarity, TINA denotes the fixed-point-only method introduced in our conference version, while TINA w/ FMI additionally applies forward marginal initialization without marginal energy regularization.
We contrast pretrained Stable Diffusion v1.4, which contains the target knowledge, with a randomly initialized UNet, which contains no learned target knowledge.
Figure~\ref{fig:mechanism_evidence}(a) shows the early trajectory on pretrained Stable Diffusion v1.4.
Compared with DDIM inversion and TINA, adding FMI places the optimization at a substantially better aligned starting point, close to the expected marginal energy progression.
TINA+ closely overlaps with TINA w/ FMI in this setting, indicating that the energy constraint remains inactive when the subsequent trajectory is already plausible rather than indiscriminately perturbing a valid inversion path.
}

\xql{
The Random UNet negative control separates initialization quality from trajectory validity more clearly.
As shown in Figure~\ref{fig:mechanism_evidence}(b), FMI again starts near the marginal reference, but TINA w/ FMI subsequently develops a severe energy deficit.
The full TINA+ objective instead intervenes during optimization and substantially mitigates this collapse.
This distinction also explains why energy validity cannot be reduced to a post-hoc rejection rule.
A validity score could identify the collapsed Random-UNet trajectory after inversion, but the optimization would still return a visually recognizable false reconstruction, yielding poor qualitative evidence of failure.
By incorporating the energy constraint into the optimization itself, TINA+ changes the discovered trajectory and turns the invalid reconstruction into an explicit generation failure.
}

\begin{table*}[t]\small
  \centering
  \color{qianlong}
  \caption{\xql{False-positive suppression throughout the progression from DDIM inversion to TINA+ on the Van Gogh and Taylor Swift benchmarks. The randomly initialized UNet contains no learned target knowledge and therefore serves as a negative control: higher LPIPS and DreamSim, lower DINO, and lower ASR indicate stronger suppression of spurious target reconstruction.}}
  \label{tab:diffusion_consistency_ablation}%
  \setlength{\tabcolsep}{4pt}
  \begin{tabular*}{\textwidth}{@{\extracolsep{\fill}}lcccccccc@{}}
    \toprule
    & \multicolumn{4}{c}{\textbf{Van Gogh}}
    & \multicolumn{4}{c}{\textbf{Taylor Swift}} \\
    \cmidrule(lr){2-5}\cmidrule(lr){6-9}
    \textbf{Method}
    & \textbf{LPIPS $\uparrow$} & \textbf{DreamSim $\uparrow$} & \textbf{DINO $\downarrow$} & \textbf{ASR $\downarrow$}
    & \textbf{LPIPS $\uparrow$} & \textbf{DreamSim $\uparrow$} & \textbf{DINO $\downarrow$} & \textbf{ASR $\downarrow$} \\
    \midrule
    DDIM
    & 0.396 & 0.212 & 0.725 & 48.0
    & 0.286 & 0.164 & 0.695 & 86.0 \\
    TINA
    & 0.361 & 0.181 & 0.769 & 60.0
    & 0.245 & 0.140 & 0.730 & 88.0 \\
    \xql{TINA w/FMI}
    & 0.482 & 0.270 & 0.633 & 28.0
    & 0.484 & 0.256 & 0.588 & 60.0 \\
    \midrule
    \xql{\textbf{TINA+}}
    & \xql{\textbf{0.651}} & \xql{\textbf{0.493}} & \xql{\textbf{0.434}} & \xql{\textbf{0.0}}
    & \xql{\textbf{0.749}} & \xql{\textbf{0.625}} & \xql{\textbf{0.400}} & \xql{\textbf{0.0}} \\
    \bottomrule
  \end{tabular*}%
\end{table*}%

\begin{figure}[t]
  \centering
  \includegraphics[width=\linewidth]{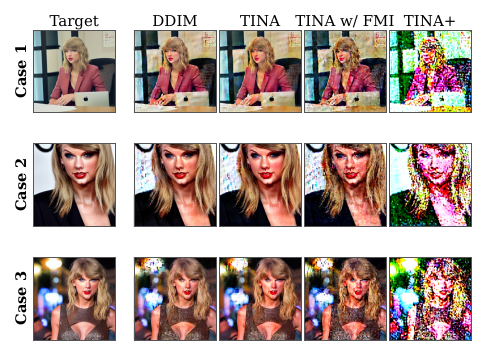}
  \caption{\xql{Qualitative false-positive reconstruction on the Taylor Swift benchmark using a randomly initialized UNet. Although the model contains no learned target knowledge, DDIM inversion and unconstrained TINA closely reconstruct the targets, while FMI alone only partially disrupts these spurious results. Marginal energy regularization substantially degrades the diffusion-inconsistent reconstructions, causing all three TINA+ outputs to fail the identity detector.}}
  \label{fig:false_positive_qualitative}
\end{figure}

\xql{
Table~\ref{tab:diffusion_consistency_ablation} and Figure~\ref{fig:false_positive_qualitative} provide complementary quantitative and qualitative evidence of false-positive suppression.
Because the randomly initialized UNet contains no target knowledge, any recognizable target reconstruction constitutes a false positive rather than evidence of successful probing.
As illustrated by the three Taylor Swift cases in Figure~\ref{fig:false_positive_qualitative}, DDIM inversion and TINA nevertheless reconstruct the target identity and scene surprisingly well, while TINA w/ FMI retains recognizable target characteristics despite introducing visible degradation.
These examples show why visual reconstruction alone cannot establish that a model retains the target knowledge, and why an aligned initialization by itself does not guarantee a valid trajectory.
In contrast, TINA+ substantially degrades the target-like reconstructions on the negative control, causing them to fail the identity detector.
This qualitative transition is consistent with the 50-target results in Table~\ref{tab:diffusion_consistency_ablation}.
On the Taylor Swift benchmark, TINA+ increases LPIPS and DreamSim to $0.749$ and $0.625$, respectively, reduces DINO similarity to $0.400$, and lowers ASR to $0.0\%$.
The same pattern holds for Van Gogh, where TINA+ increases LPIPS and DreamSim to $0.651$ and $0.493$, respectively, reduces DINO similarity to $0.434$, and lowers the false-positive ASR to $0.0\%$.
Together with Figure~\ref{fig:mechanism_evidence}, these results show that FMI supplies an aligned starting point, whereas marginal energy regularization is necessary to prevent subsequent optimization from exploiting diffusion-inconsistent paths.
}

\subsection{Latent Embedding Analysis}
\begin{figure}[t]
    \centering
    \includegraphics[width=\linewidth]{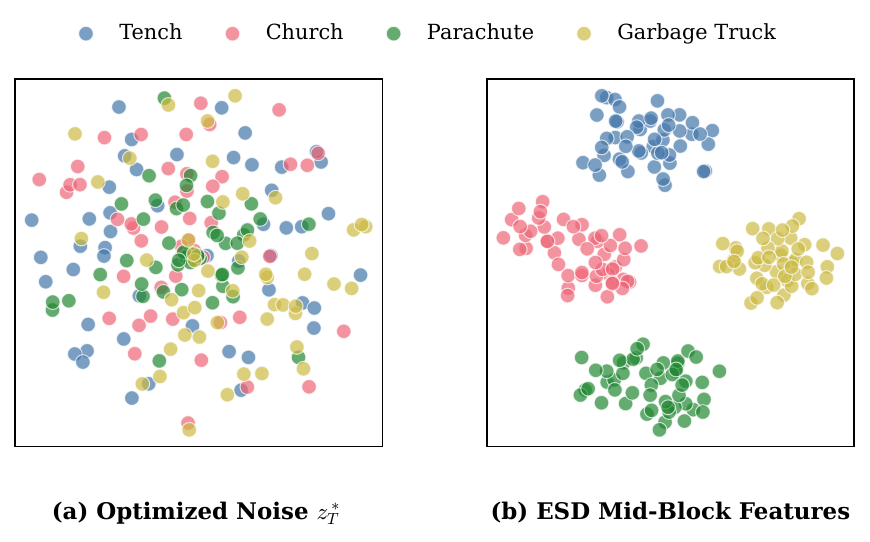}
    \caption{t-SNE visualization of (a) the optimized initial noises $z_T^*$ produced by \xql{TINA+} and (b) their corresponding deep ESD UNet activations, extracted from the last ResNet block of \texttt{mid\_block}, for four erased concepts. Each concept contains 50 target images.}
    \label{fig:tsne}
\end{figure}

\xql{Having examined the diffusion consistency of the discovered trajectories, we further investigate the internal responses that they elicit in the corresponding erased models.}
\xql{For four erased concepts (Tench, Church, Parachute, and Garbage Truck), we collect 50 initial noises $z_T^*$ optimized by TINA+ against each corresponding ESD-sanitized model and visualize both the noises and their induced UNet activations using t-SNE.}
We then feed these optimized noises into their corresponding ESD UNets under a null-text condition and extract the intermediate representations through a forward inference.
For visualization, we apply t-SNE (perplexity 30, with PCA initialization) to both the flattened noise vectors and the global-average-pooled features extracted from the last ResNet block of \texttt{mid\_block}.
As shown in Figure~\ref{fig:tsne}, the optimized noises themselves exhibit little concept-wise separation in the latent space, consistent with their noise-like nature.
However, their corresponding internal UNet activations become clearly separable by concept, forming distinct clusters.
\xql{This result shows that although the optimized $z_T^*$ remains unstructured in the input latent space, concept-wise structure emerges in the internal responses of the corresponding erased models. Together with the reconstruction results, this observation is consistent with TINA+ activating residual concept-related representations through text-free trajectories.}

\subsection{Generalization to DiT-Based Architectures}
\label{sec:dit_generalization}

\xql{While the main experiments focus on UNet-based diffusion models (e.g., Stable Diffusion v1.4), an important question is whether TINA+ generalizes to fundamentally different model architectures. To investigate this, we evaluate TINA+ on \texttt{PixArt-XL-2-512x512}~\cite{chen2023pixart}, a text-to-image model built on the Diffusion Transformer (DiT) architecture~\cite{peebles2023scalable} rather than a UNet backbone.}

\xql{We conduct this evaluation on the ``Parachute'' concept.}
\xql{Additionally, as there are no publicly available erased versions of this DiT model, we first apply the ESD~\cite{gandikota2023erasing} method to erase the parachute concept from the model, and then perform our TINA+ attack on the resulting erased model.}

\begin{figure}[t]
    \centering
    \includegraphics[width=\linewidth]{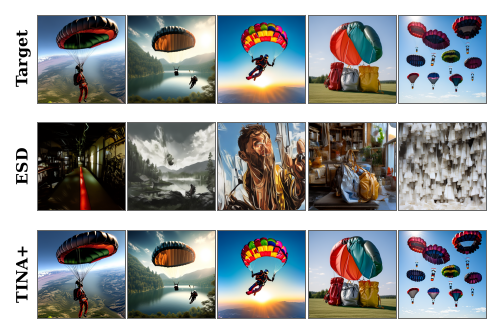}
    \caption{\xql{Generalization of TINA+ to a DiT-based architecture (\texttt{PixArt-XL-2-512x512}). Each column shows a target image, the ordinary output of the ESD-erased model, and the corresponding TINA+ reconstruction. ESD removes the Parachute concept under normal sampling, whereas TINA+ recovers diverse target-specific parachutes.}}
    \label{fig:dit_generalization}
\end{figure}

\xql{Across 50 target images, the erased DiT model obtains a $0.0\%$ ASR under normal prompt-based sampling, whereas TINA+ successfully recovers the Parachute concept in 42 cases, achieving an $84.0\%$ ASR.}
\xql{As shown in Figure~\ref{fig:dit_generalization}, the ordinary ESD outputs lose the target concept and scene content, while TINA+ reconstructs diverse target-specific parachutes from the same erased DiT model.}
\xql{These results demonstrate that the vulnerability exposed by TINA+ is not limited to UNet-based models, but also extends to DiT-based models.}

\section{Conclusion}
\xql{We exposed a fundamental limitation of the text-centric concept erasure paradigm: severing text-image links is insufficient to truly remove the underlying visual knowledge.}
\xql{We introduced TINA+, a diffusion-consistent text-free attack that probes residual visual knowledge while bypassing textual controls entirely.}
\xql{TINA+ discovers text-free generative trajectories while suppressing spurious inversion paths that are inconsistent with the diffusion process.}
\xql{Extensive experiments across diverse concepts, erasure methods, and different model architectures demonstrate that TINA+ can reliably expose residual visual knowledge in concept-erased models.}
\xql{These findings indicate that existing methods may obscure visual knowledge by disrupting its textual associations rather than fully removing it, motivating future unlearning techniques that operate directly on internal visual representations.}

\bibliographystyle{IEEEtran}
\bibliography{IEEEabrv,main}

\begin{IEEEbiography}
[{\includegraphics[width=1in,height=1.25in,clip,keepaspectratio]{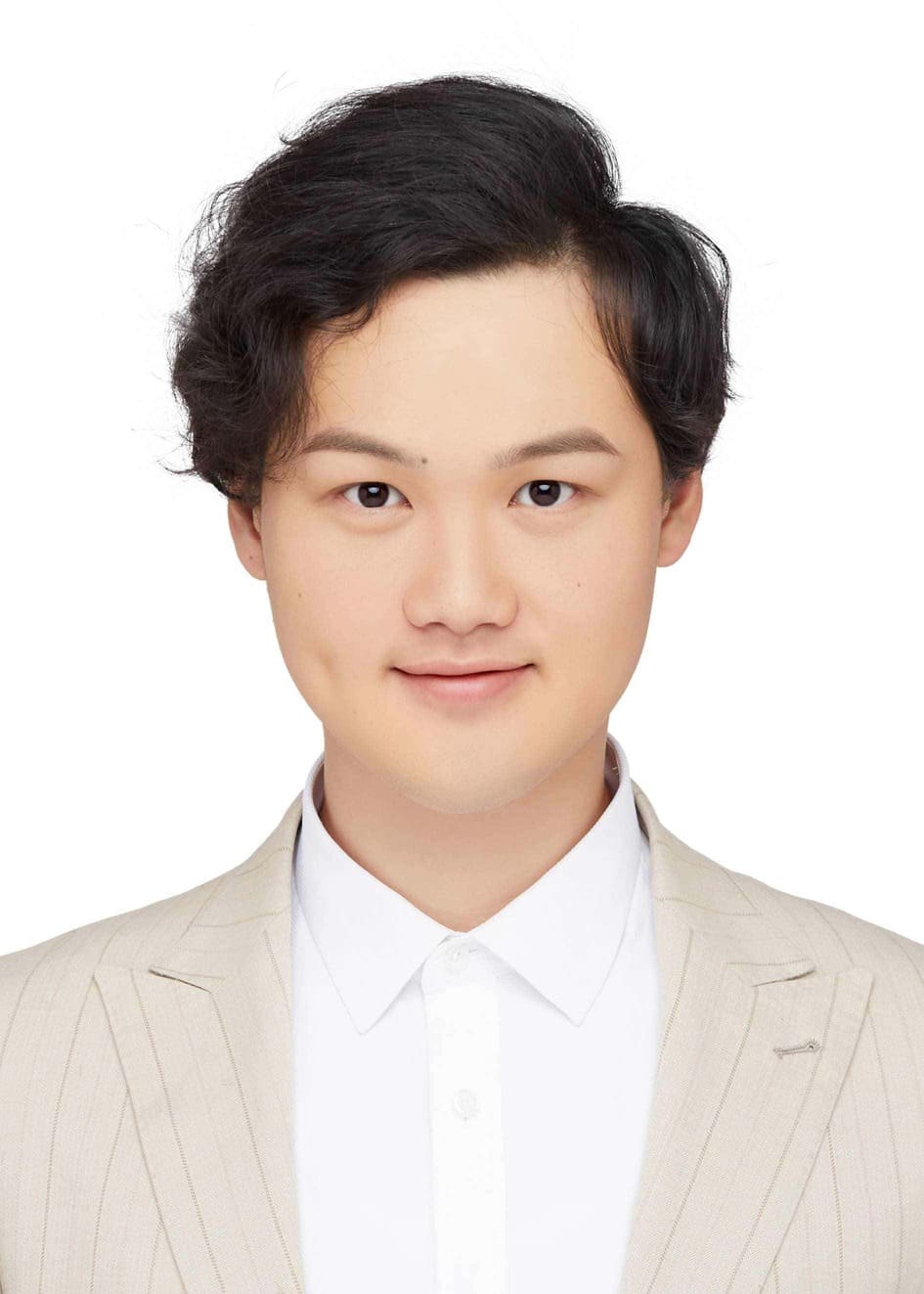}}]{Qianlong Xiang}
received the B.E. degree from the School of Computer Science and Technology, Harbin Institute of Technology, Weihai, China, in 2022. He is currently pursuing the Ph.D. degree at the School of Computer Science and Technology, Harbin Institute of Technology, Shenzhen, China. His research has been published in top-tier conferences including CVPR. He has served as a reviewer for various conferences and journals, such as IEEE TPAMI, ACM MM, and IEEE TCSVT. His main research interests include model compression and generative AI.
\end{IEEEbiography}

\begin{IEEEbiography}[{\includegraphics[width=1in,height=1.25in,clip,keepaspectratio]{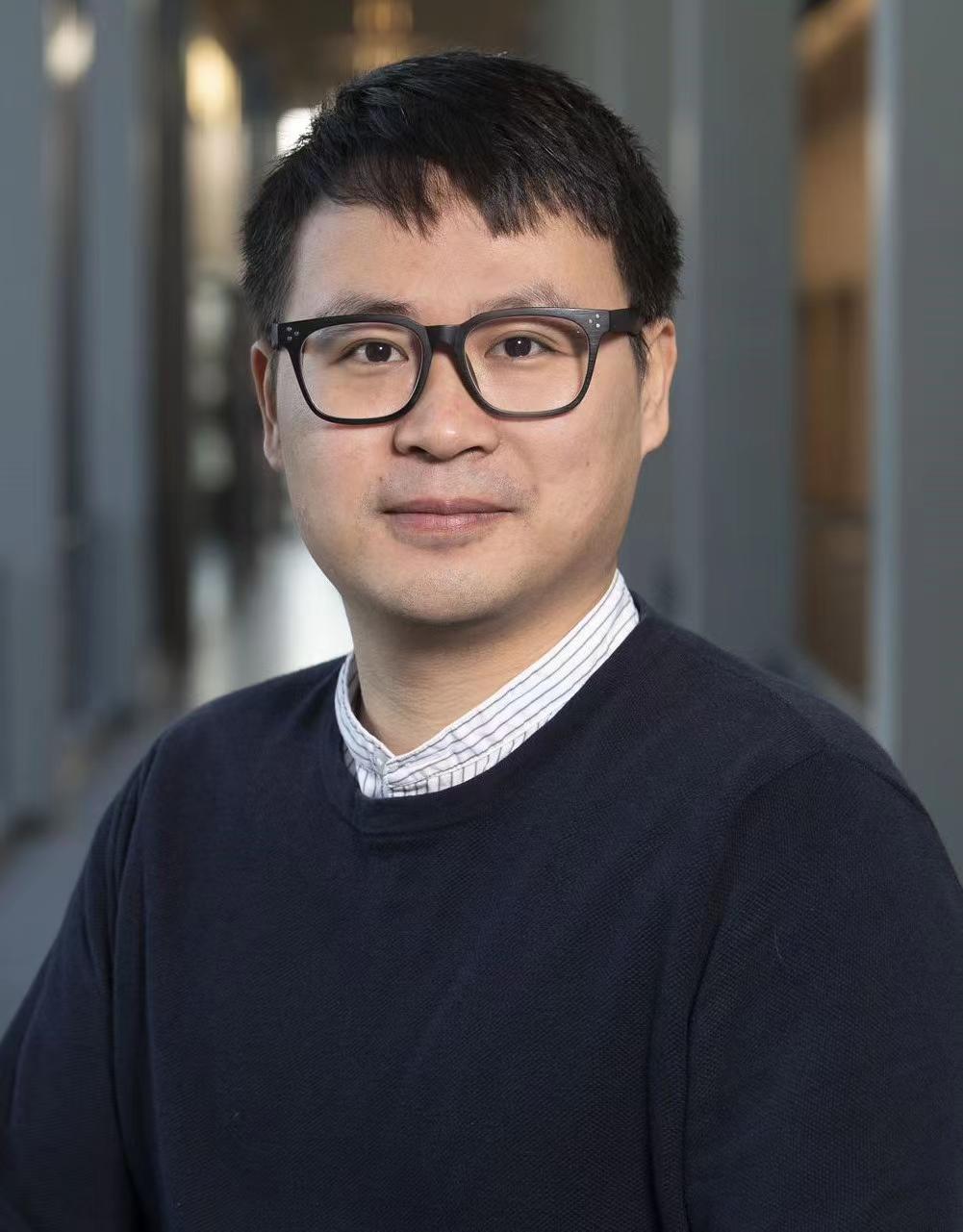}}]{Miao Zhang}
(Member, IEEE) received the Ph.D. degree from the University of Technology Sydney (UTS), Australia. He is currently a Professor with Harbin Institute of Technology (Shenzhen), Shenzhen, China. Before that, he was an Assistant Professor at Aalborg University, Aalborg, Denmark. His primary research interests include AutoML, model compression, efficient learning, and continual learning.
\end{IEEEbiography}

\begin{IEEEbiography}
[{\includegraphics[width=1in,height=1.25in,clip,keepaspectratio]{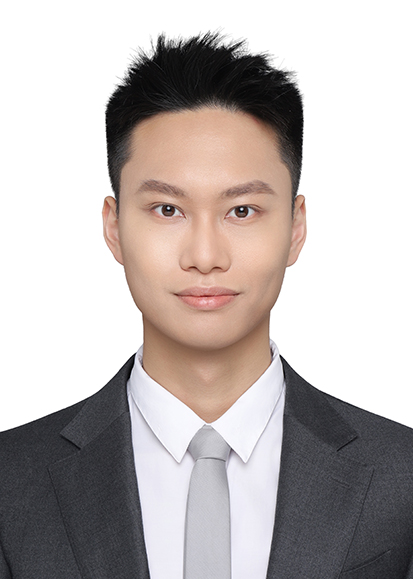}}]{Kun Wang}
is a Research Fellow at the National University of Singapore. He received the Ph.D. degree from the School of Software, Shandong University, Jinan, China, in 2026, and the B.E. degree from the School of Computer Science and Technology, Shandong University, Qingdao, China, in 2022. His research has been published in top-tier conferences and journals, including ACM MM, IEEE TIP, and ACM TOIS. He also serves as a reviewer for IEEE TPAMI, IEEE TIP, IEEE TKDE, and IEEE TMM. His main research interests include information retrieval and video analysis.
\end{IEEEbiography}

\begin{IEEEbiography}[{\includegraphics[width=1in,height=1.25in,clip,keepaspectratio]{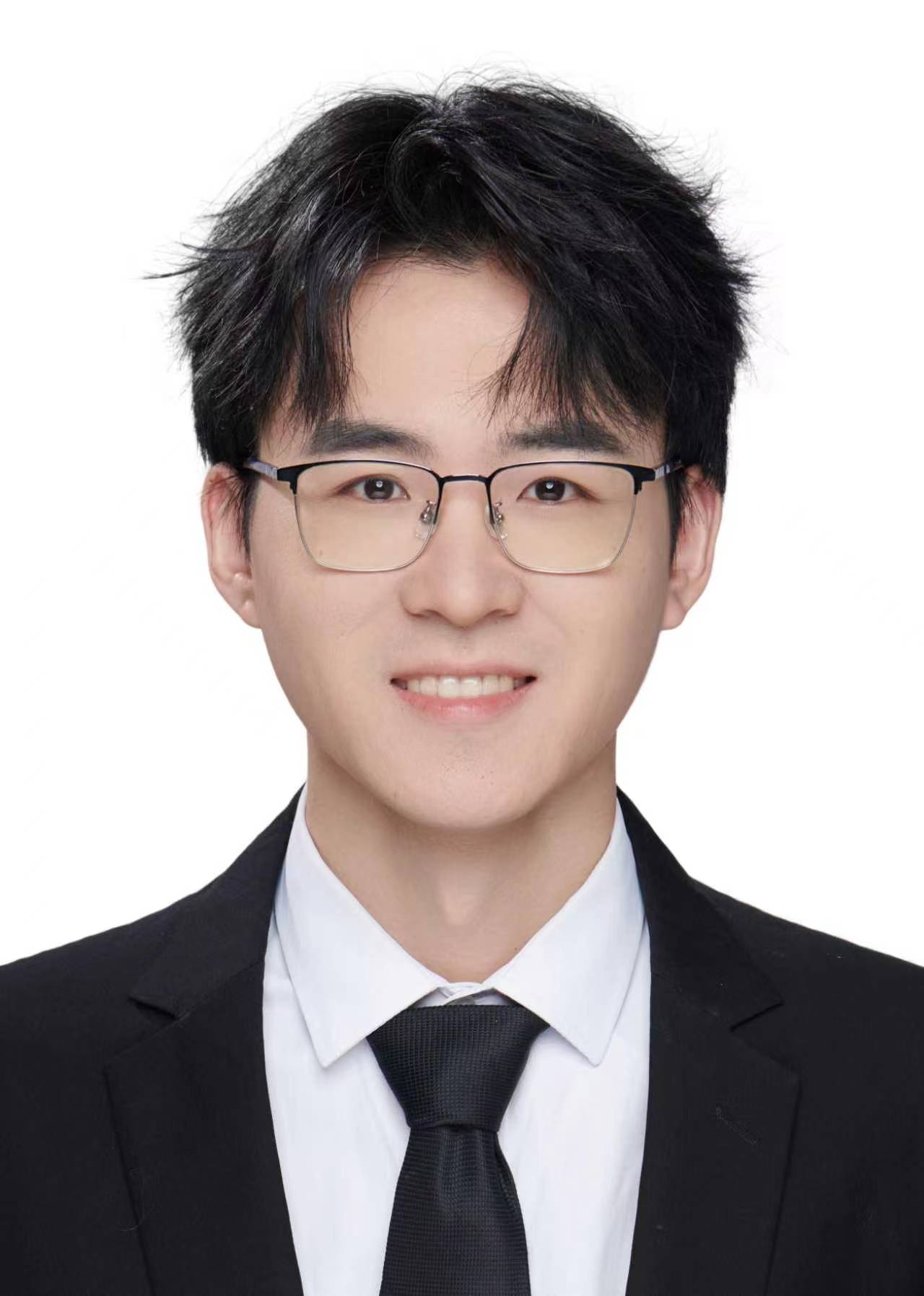}}]{Haoyu Zhang}
received the M.S. degree from Shandong University, in 2023. He is currently working toward the doctor's degree with the School of Computer Science and Technology, Harbin Institute of Technology (Shenzhen). His research interests include egocentric vision and spatial understanding. His work has been published in several top-tier conferences and journals, including IEEE TPAMI, CVPR, NeurIPS, ICML, AAAI, and ACM MM. He has served as a Reviewer for various conferences and journals, such as CVPR, ICCV, NeurIPS, ICML, ICLR, IEEE TPAMI, IEEE TKDE, and IEEE TMM.
\end{IEEEbiography}

\begin{IEEEbiography}[{\includegraphics[width=1in,height=1.25in,clip,keepaspectratio]{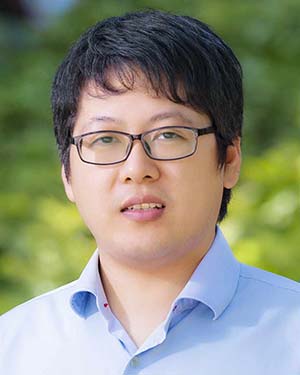}}]{Junhui Hou}
(Senior Member, IEEE) is a Professor with the Department of Computer Science, City University of Hong Kong (CityUHK). His research interests include multidimensional visual computing. He received the Early Career Award and Research Fellow from the Hong Kong Research Grants Council,  the Excellent Young Scientists Fund from NSFC, the IEEE TIP Best Paper Award, and the CityUHK Presidential Research Excellence Award for Junior Faculty. He is serving as a Senior Area Editor for IEEE TIP and an Associate Editor for IEEE TVCG and TMM. He served as an Associate Editor for IEEE TIP and TCSVT.
\end{IEEEbiography}

\begin{IEEEbiography}[{\includegraphics[width=1in,height=1.25in,clip,keepaspectratio]{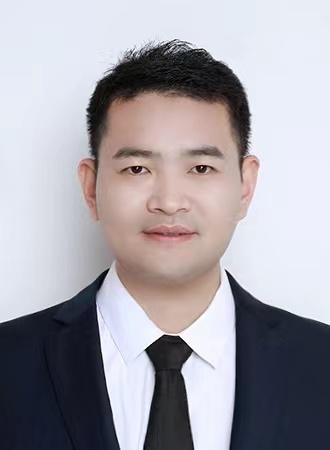}}]{Liqiang Nie}
(Senior Member, IEEE) received the B.Eng. degree from Xi'an Jiaotong University, Xi'an, China, and the Ph.D. degree from the National University of Singapore, Singapore. Currently, he is a Professor with the School of Computer Science and Technology, Harbin Institute of Technology, Shenzhen, China, and a Fellow of the IAPR.
Dr. Nie serves as an Associate Editor for IEEE TIP, IEEE TKDE, IEEE TMM, IEEE TCSVT, ACM ToMM, and Information Sciences. Meanwhile, he is the chair of ICMR 2025, ICME 2025, and ACM MM 2027, and a member of the ICME steering committee.
His main interests are multimedia content analysis and information retrieval. He has published over 150 papers and 5 books in these fields, with more than 30,000 Google Scholar citations.
He has received many awards over the past four years, including SIGMM Rising Star in 2020, MIT TR35 China 2020, SIGIR Best Student Paper in 2021, IEEE AI's 10 to Watch in 2022, ACM MM Best Paper Award in 2022, and the National Youth Science and Technology Award in 2024.
\end{IEEEbiography}

\end{document}